\documentclass[lettersize,journal]{IEEEtran}
\usepackage{amsmath,amsfonts}
\usepackage{algpseudocode}
\usepackage{algorithmicx}
\usepackage{algorithm}
\usepackage{array}
\usepackage[caption=false,font=normalsize,labelfont=sf,textfont=sf]{subfig}
\usepackage{textcomp}
\usepackage{stfloats}
\usepackage{url}
\usepackage{verbatim}
\usepackage{graphicx}
\usepackage{cite}
\usepackage[utf8]{inputenc} 
\usepackage[T1]{fontenc}    
\usepackage{hyperref}       
\usepackage{url}            
\usepackage{booktabs}       
\usepackage{amsfonts}       
\usepackage{nicefrac}       
\usepackage{microtype}      
\usepackage{xcolor}         
\definecolor{myred}{RGB}{174, 46, 46}   
\definecolor{myblue}{RGB}{28, 134, 167} 
\definecolor{mygrey}{RGB}{137, 138, 140}
\definecolor{mydarkred}{RGB}{173, 45, 45}
\definecolor{mydarkgreen}{RGB}{15, 163, 25}
\definecolor{supp_blue}{RGB}{27, 161, 226}
\definecolor{supp_green}{RGB}{96, 169, 23}

\usepackage{pifont} 
\usepackage{arydshln}
\usepackage{mathtools}
\usepackage{amsmath}
\usepackage{bm}
\usepackage{graphicx}
\usepackage{multirow}
\usepackage{wrapfig}
\usepackage{afterpage}
\usepackage{algpseudocode}
\definecolor{spc}{RGB}{119, 107, 170}
\definecolor{pct}{rgb}{0.7, 0, 0.2}
\newcommand{\spc}{\textcolor{spc}{$\mathbf{\circ}$\,}}
\newcommand{\pct}{\textcolor{pct}{$\bullet$\,}}
\usepackage{fontawesome5}
\hypersetup{
    colorlinks=true,
    linkcolor=red,
    citecolor=cyan,
    filecolor=magenta,      
    urlcolor=magenta,
}
\usepackage{tikz}
\usepackage{graphicx}
\usepackage{capt-of}

\usepackage{multirow}
\usepackage{amssymb}
\usepackage{pifont}
\usepackage{enumitem}
\let\oldding\ding
\renewcommand{\ding}[2][1]{\scalebox{#1}{\oldding{#2}}}

\begin{document}

\title{Post-training Quantization for Hybrid Iterative Generative Models}

\author{Jing~Gao,
        Junyi~Wu,
        Wei~Wang,~\IEEEmembership{Senior Member,~IEEE,}
        Yan~Yan,~\IEEEmembership{Senior Member,~IEEE,}
        Yao~Zhao,~\IEEEmembership{Fellow,~IEEE.}
\IEEEcompsocitemizethanks{%
  \IEEEcompsocthanksitem Jing Gao, Wei Wang, and Yao Zhao are with the Institute of Information Science, Beijing Jiaotong University, Beijing, China. E-mail: \{22331159, wei.wang, yzhao\}@bjtu.edu.cn.
  (Corresponding author: Wei Wang.)%
  \IEEEcompsocthanksitem Junyi Wu and Yan Yan are with the University of Illinois Chicago, Chicago, IL, USA. E-mail: \{jwu834, yyan55\}@uic.edu.%
}

\thanks{Manuscript ID: TCSVT-27936-2026.

Digital Object Identifier 10.1109/TCSVT.2026.3710967}
}

\markboth{IEEE Transactions on Circuits and Systems for Video Technology, Manuscript ID: TCSVT-27936-2026}%
{Gao \MakeLowercase{\textit{et al.}}: Post-training Quantization for Hybrid Iterative Generative Models}

\IEEEpubid{\begin{minipage}{\textwidth}\ \centering
        \textcopyright{} 2026 IEEE. 
        Personal use of this material is permitted. \\
        However, permission to use this material for any other purposes must be obtained from the IEEE by sending an email to pubs-permissions@ieee.org.
\end{minipage}}

\maketitle

\begin{abstract}
Iterative Generative Models (IGMs) span autoregressive and diffusion paradigms, and hybrid variants that couple them can achieve remarkable image-generation fidelity.
However, their iterative inference incurs substantial computational overhead, making Post-training Quantization (PTQ) appealing for acceleration, while directly applying vanilla PTQ to hybrid IGMs can trigger model collapse.
By analyzing these failures, we identify two critical challenges:
Excessive Outliers (EOs) in the activations create an irreconcilable trade-off between preserving normal precision and covering EOs, resulting in severe degradation in generation quality; Amplified Anomalies (AAs) arising unpredictably from minor quantization errors, create a mismatch between calibration and inference, thus iteratively triggering model collapse.
To address these challenges, we introduce HyGenQ, a PTQ framework for hybrid IGMs. HyGenQ comprises Hierarchical Cluster Decoupling (HCD) and Scaling Recalibration (SR).
HCD identifies and decouples outlier channels via a multi-stage clustering process, effectively isolating EOs while maintaining normal value precision, thereby alleviating performance degradation.
SR scales AAs beyond Gaussian Bound, thereby avoiding model collapse caused by aggressive truncation.
Extensive experiments demonstrate that HyGenQ successfully quantizes representative hybrid IGMs to 8-bit precision (W8A8), significantly outperforming existing baselines and validating its robustness across different model families.

\end{abstract}

\begin{IEEEkeywords}
Iterative Generative Models, Outliers, Model Compression, Post-training Quantization, Efficiency.
\end{IEEEkeywords}

\section{Introduction}
\label{sec:introduction}
\IEEEPARstart{I}{terative} generative models (IGMs) generate images via multi-step refinement at inference time, where intermediate states are repeatedly updated until convergence.
Representative paradigms include autoregression~\cite{esser2021taming,gregor2014deep,van2016pixel,chen2018pixelsnail} and diffusion~\cite{ho2020denoising,dhariwal2021diffusion}.
Notably, hybrid IGMs~\cite{li2025autoregressive,tang2024hartefficientvisualgeneration} combine token-by-token autoregression for global structure and long-range dependencies with diffusion-style iterative refinement for local details.
This coupling leverages complementary strengths of both paradigms to enable high-fidelity image generation.

Nevertheless, hybrid IGMs typically incur substantial inference cost due to their inherently iterative nature:
autoregressive models predict tokens sequentially, while diffusion models perform step-by-step sampling~\cite{ho2020denoising,dhariwal2021diffusion}.
In practice, the dominant bottleneck often lies in matrix multiplications~\cite{esser2019learned,zhang2024vitfpga} within linear layers, prolonging inference time and hindering the feasibility of real-time applications of hybrid IGMs.

Model quantization \cite{shen2020q,10239177,9927185,11006154,10671591,11342323,10528335} has emerged as a potent compression method to reduce computational overhead and enhance inference speed.
Among various quantization approaches, Post-training Quantization (PTQ) \cite{li2023repq,wu2025ptq4dit,he2023ptqd,nagel2020up} stands out, as it eliminates the necessity for retraining \cite{he2023ptqd} and requires only a small dataset for calibration. 
These properties make PTQ especially appealing for hybrid IGMs, where generation involves many repeated steps and re-training is often costly.

\IEEEpubidadjcol

However, applying vanilla PTQ to hybrid IGMs is considerably more challenging than to standard feed-forward inference.
During iterative generation, activations can become highly non-stationary and context-dependent~\cite{shi2025meaningless}, and small quantization perturbations may propagate and accumulate across many iterations~\cite{li2023q}, potentially destabilizing the entire process.

\begin{figure*}[t]
  \centering
  \vspace*{-5mm}
  \includegraphics[width=2\columnwidth]{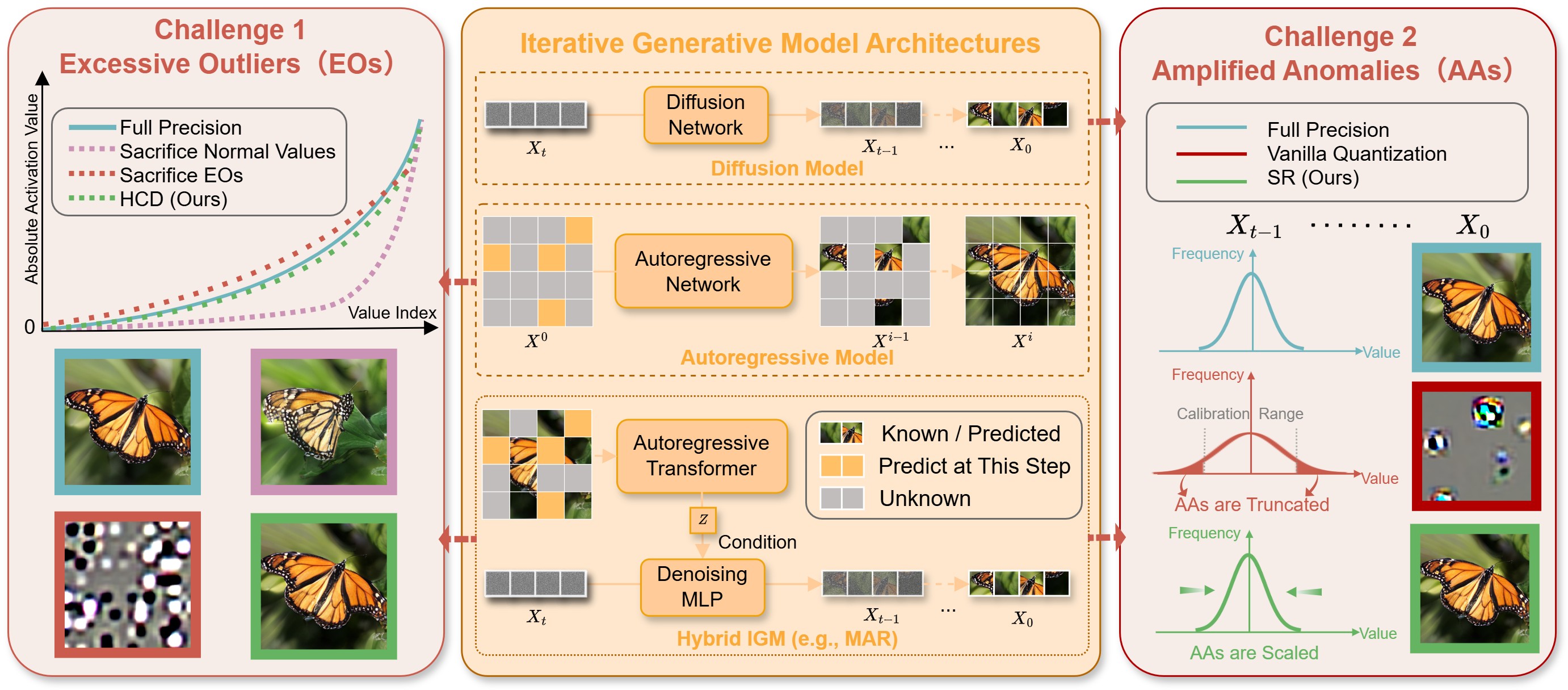}
  \vspace{-4mm}
  \caption{
  \textbf{Overview of iterative generative model paradigms and post-training quantization challenges.}
  IGMs include diffusion, autoregressive, and hybrid pipelines that couple autoregressive context updates with diffusion-style refinement (e.g., MAR, where an autoregressive Transformer conditions a denoising MLP to iteratively sample new tokens and update the token sequence).
  \textbf{(Left)}
  Excessive Outliers create an irreconcilable trade-off between sacrificing normal values and sacrificing outliers.
  Our Hierarchical Cluster Decoupling addresses this by isolating outlier channels while maintaining quantization precision for normal channels.
  \textbf{(Right)}
  Amplified Anomalies arising unpredictably from minor quantization errors, exceed vanilla PTQ's static calibration ranges during diffusion, thus causing truncation artifacts.
  Our Scaling Recalibration scales anomalies into valid ranges, avoiding model collapse.
  }
  \vspace{-4mm}
  \label{figure1}
\end{figure*}

Specifically, we observe two failure modes that hinder effective PTQ for hybrid IGMs:
\textbf{E}xcessive \textbf{O}utliers (\textbf{EOs}) in the activations pose a significant obstacle for uniform quantization, making it difficult to strike an appropriate balance between preserving the precision of the predominant normal values and maintaining the representational range for outliers, which are often critical to generation quality.
Sacrificing either component can lead to substantial information loss and even model collapse, as illustrated in Figure \ref{figure1} (Left).
\textbf{A}mplified \textbf{A}nomalies (\textbf{AAs}) in the activations originate from minor quantization errors during the diffusion procedure. These errors, once amplified, introduce unpredictable deviations that propagate into subsequent denoising steps. Vanilla PTQ methods \cite{li2023repq,xiao2023smoothquant,shang2023ptq4dm}, constrained by static calibration ranges, are incapable of handling such unpredictable overflows during inference. Consequently, they truncate these critical anomalies, ultimately leading to severe model collapse, as shown in Figure \ref{figure1} (Right).

To address these challenges, we propose \textbf{HyGenQ}, a post-training quantization framework for hybrid IGMs. HyGenQ integrates two modules:
To address the issue of EOs, we propose \textbf{H}ierarchical \textbf{C}luster \textbf{D}ecoupling (\textbf{HCD}).
HCD is motivated by the observation that channels exhibiting EOs remain highly consistent across iterations within the same layer.
Leveraging this consistency, HCD aims to detect and isolate outlier channels, thereby enabling the separation of outliers and normal values to better preserve generation quality.
Specifically, HCD adopts a three-stage hierarchical clustering process: it first partitions channels into multiple fine-grained clusters, then aggregates them into a main cluster and a non-main cluster based on a Quantization Error Metric, and finally decouples normal and outlier channels accordingly for independent quantization.
Our HCD facilitates a favorable trade-off between retaining outliers and preserving the precision of normal values, with minimal compromise in overall quantization accuracy.

To further address the issue of AAs, we propose \textbf{S}caling \textbf{R}ecalibration (\textbf{SR}).
SR is based on the observation that anomalies are non-stationary and irregular, and typically occur during the early denoising process.
Leveraging this property, SR defines a fixed Gaussian Bound based on the activation distribution of the Full-Precision model and dynamically monitors activation fluctuation across channels. Channels that exceed this bound are scaled accordingly to restore stability.
SR effectively mitigates the risk of model collapse caused by anomalies, while preserving the overall generation fidelity.

Our \textbf{HyGenQ} successfully quantizes representative IGMs to 8-bit precision (W8A8), corroborating its robustness and generality.
In particular, for hybrid IGMs, even when disregarding the severe collapse risks induced by AAs, existing PTQ methods still suffer from substantial degradation in generation quality. In contrast, HyGenQ consistently outperforms these baselines by a large margin.
More critically, when all linear layers are quantized, mainstream PTQ methods often result in complete model collapse, whereas HyGenQ prevents such catastrophes and revives generation capability.

\section{Backgrounds and Related Works}
\label{sec:related Works}
\subsection{Iterative Generative Models}\label{sec:MAR}
Iterative generative models (IGMs) synthesize images through multi-step inference, where intermediate states are repeatedly updated until convergence.
Representative paradigms include diffusion models that generate samples via iterative denoising~\cite{ho2020denoising,dhariwal2021diffusion,nichol2021improved},
autoregressive models that predict discrete visual tokens sequentially~\cite{esser2021taming,chen2018pixelsnail,sun2024autoregressive},
and hybrid architectures that couple token-wise autoregressive modeling with diffusion-style iterative refinement~\cite{li2025autoregressive,tang2024hartefficientvisualgeneration}.
In hybrid IGMs, the tight coupling of two paradigms leads to highly non-stationary activations, posing severe challenges for post-training quantization. In this work, we use the Masked Autoregressive (MAR) model~\cite{li2025autoregressive} as a representative hybrid pipeline and briefly review its architecture and inference procedure.

Concretely, for a sequence of tokens \(\{x_1, x_2, \dots, x_n\}\), where the subscript \(1 \leq i \leq n\) denotes the sequence order, the autoregressive module (typically implemented via Transformers \cite{vaswani2017attention}) produces a vector \(z_i=f(\cdot)\) based on the previous tokens \(\{x_1, x_2, \dots, x_{i-1}\}\), which serves as the condition \cite{zhang2023adding,choi2021ilvr} for the diffusion network (a small MLP) to model the probability distributions for the subsequent tokens.

To generate samples from the distribution, MAR models apply a reverse diffusion process of iterative denoising following the common practice of DDPM \cite{ho2020denoising,dhariwal2021diffusion}:
\begin{equation}
x_{t-1}=\frac{1}{\sqrt{\alpha_t}}\Big(x_t-\gamma_t\,\varepsilon_\theta(x_t\vert t,z)\Big)+\sigma_t\delta,
\label{eq:denoising}
\end{equation}
where  the factor \( \gamma_t \triangleq \frac{1 - \alpha_t}{\sqrt{1 - \bar{\alpha}_t}} \) controls the influence of predicted noise,  
\( \varepsilon_\theta(x_t \vert t, z) \) denotes the predicted noise based on \(x_t\).

For training, MAR adopts Diffusion Loss (DiffLoss)~\cite{tan2024diffloss} to learn the distribution $p(x\vert z)$, defined as follows~\cite{ho2020denoising,nichol2021improved,dhariwal2021diffusion}:
\begin{equation}
\mathcal{L}(z, x) = \mathbb{E}_{\varepsilon, t}\!\left[\,\left\|\,\varepsilon - \varepsilon_{\theta}(x_t \vert t, z)\,\right\|^{2}\right].
\label{eq:diff_loss}
\end{equation}

Gradients from DiffLoss propagate through \(z\) into the autoregressive backbone, thereby enabling end-to-end joint training.
Moreover, the same \(z\) is reused for multiple timesteps \(t\) to better utilize the loss without recomputing \(z\) in practice.

At inference, with bidirectional attention~\cite{he2022masked} enabling each step to generate a random subset of tokens~\cite{chang2022maskgit}, the procedure progressively refines the sequence until the full image is synthesized.
Overall, MAR can be summarized as \textit{modeling the interdependence of tokens by autoregression, jointly with the per-token distribution by diffusion}~\cite{li2025autoregressive}.

However, the iterative inference shared by IGMs often incurs substantial computational overhead, motivating further research on efficient deployment and acceleration in practical settings.

\subsection{Post-training Quantization}\label{sec:PTQ}

Model quantization \cite{he2023ptqd,li2023repq,wei2022qdrop,yuan2022ptq4vit,shen2020q}, which transforms full precision parameters into low-bit integers\cite{806625}, can substantially decrease storage needs and computational complexity, thereby enhancing inference speed in practice.
Formally, post-training quantization calibrates the quantization parameters on a small set of representative samples, and these parameters are then used for low-bit computation during inference~\cite{yuan2022ptq4vit,wei2022qdrop}.
For an input activation tensor $x$ of a linear layer collected during calibration, the affine uniform quantization process can be expressed as follows:

\begin{equation}
\begin{split}
Q(x) &= \text{clip}\left( \left\lfloor \frac{x}{\lambda} \right\rceil + \beta, \; 0, \; 2^b - 1 \right), \\
\quad \text{where } \lambda &= \frac{\max(x) - \min(x)}{2^b - 1}, \; \beta = - \left\lfloor \frac{\min(x)}{\lambda} \right\rceil.
\end{split}
\label{quant_all}
\end{equation}
Here, \(x\) denotes the full-precision tensor, \( \lfloor \cdot \rceil \) signifies the round function \cite{nagel2020up}, $\lambda$ and $\beta$ denote the quantization scale and zero-point, respectively.
The formula maps \(x\) to the integer range \([0, 2^b - 1]\), where \(b\) indicates the quantization bit-width.
Among the numerous quantization methods, Post-training Quantization (PTQ) stands out as an effective model compression strategy.
It is particularly advantageous in situations calling for swift deployment and enhanced inference speed, as it eliminates the necessity for retraining \cite{he2023ptqd,liu2023pd,wei2022qdrop}.

In recent years, PTQ methods have demonstrated remarkable success across various models, including Vision Transformers (ViTs) \cite{wei2022qdrop,yuan2022ptq4vit,liu2023pd,li2023repq,li2022q,806625} and Diffusion models \cite{he2023ptqd,li2023q,wang2023towards,sun2024tmpq,huang2024tfmq}.
However, those PTQ methods that excel within their fields fail to yield satisfactory outcomes when applied to hybrid IGMs due to the token-by-token modeling and multiple iterations inherent in hybrid IGMs.
The application of PTQ to hybrid IGMs remains an area that has yet to be thoroughly explored.

\section{Quantization Challenges in Hybrid IGMs}
\label{observations and analyses}

Different from conventional ViTs~\cite{dosovitskiy2020image,vaswani2017attention} and diffusion models~\cite{ho2020denoising,dhariwal2021diffusion}, hybrid IGMs jointly model token interdependence by autoregression and the distribution of each token by diffusion. In this coupled process, autoregressive updates provide evolving contextual conditions for diffusion sampling, while the sampled tokens are further incorporated into subsequent contexts~\cite{li2025autoregressive}. Such repeated conditional interactions make activations more sensitive to quantization perturbations, leading to more unstable and extreme activation distributions.

Under quantization, this heightened sensitivity manifests as two abnormal activations: Excessive Outliers and Amplified Anomalies.
Excessive Outliers refer to unusually large activations that become more pronounced through the interaction between autoregressive context modeling and diffusion sampling.
Amplified Anomalies arise after quantization perturbations enter the iterative denoising process, where minor rounding or clipping errors can be amplified through diffusion sampling and produce activations beyond the calibration range.

As the coupled generation process proceeds, these abnormal activations are progressively accumulated and amplified, eventually driving the quantized model away from the full-precision inference path and causing severe model collapse. 
We next use Masked Autoregressive model~\cite{li2025autoregressive} as a representative hybrid IGM to analyze Excessive Outliers (Section~\ref{sec:EOs}) and Amplified Anomalies (Section~\ref{sec:AAs}) in detail.

\subsection{Excessive Outliers}\label{sec:EOs}

\begin{figure}[t]
  \centering
  \vspace*{-7mm}
  \includegraphics[width=\linewidth]{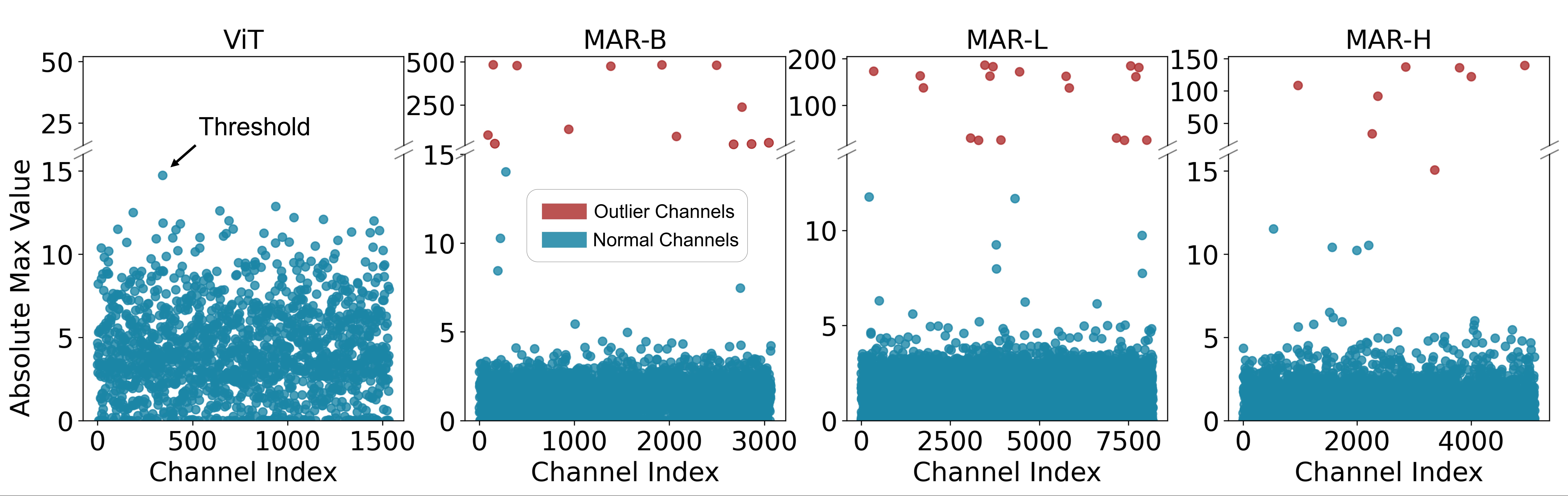}
  \vspace{-8mm}
  \caption{
  Activation distribution comparison in a linear layer between hybrid IGMs and a conventional ViT-B. 
  Using the maximum activation in ViT-B as a reference threshold, we highlight channels in hybrid IGMs that exceed this range. 
  Hybrid IGMs exhibit sparse but significantly more pronounced outliers, revealing a more severe imbalance across channels.
  }
  \vspace{-2mm}
  \label{figure_comparison}
\end{figure}

\begin{figure}[t]
  \centering
  \vspace*{-2mm}
  \includegraphics[width=\columnwidth]{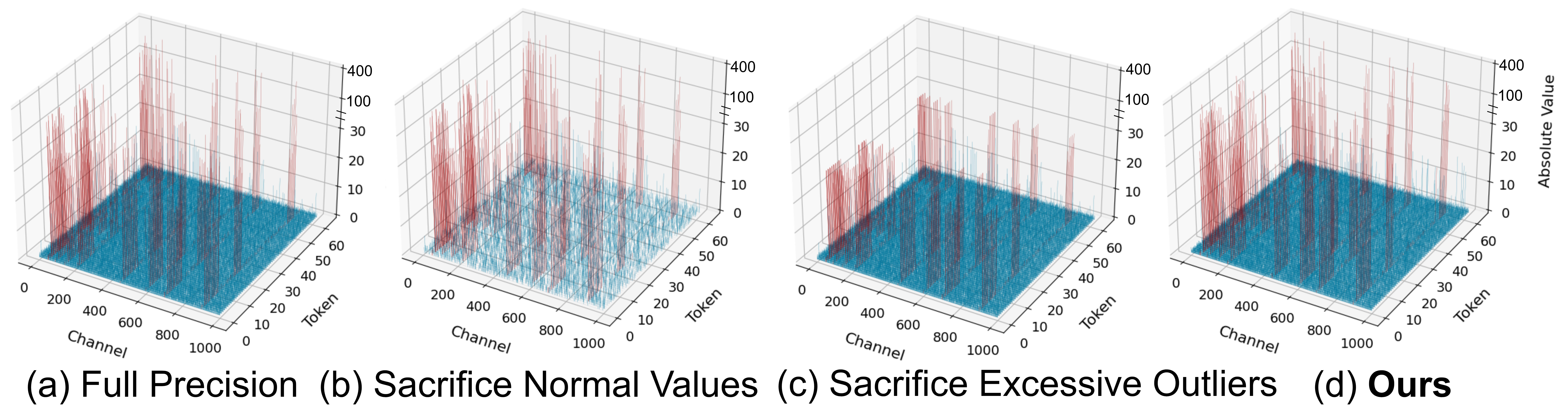}
  \vspace{-7mm}
  \caption{Comparison of full-precision and differently quantized activations in a linear layer, each including \textcolor{myred}{\textbf{Excessive Outliers (EOs)}} and \textcolor{myblue}{\textbf{Normal Values}}.
  Notably, EOs significantly exceed the range of normal values.
  \textbf{(a)} Full-Precision: Original distribution of activations;
  \textbf{(b)} Sacrifice Normal Values: Preserve all outliers while compressing most normal values;
  \textbf{(c)} Sacrifice Excessive Outliers: Preserve normal values while truncating outliers with MSE metric;
  \textbf{(d)} Ours: Our method isolates outlier channels to preserve the range of outliers and maintain quantization precision for normal values.
  }
  \vspace{-5mm}
  \label{figure2}
\end{figure}

We observe excessive outliers in hybrid IGM activations, which make post-training quantization difficult. We analyze this phenomenon from the following aspects.

\textbf{(i) Stepwise generation under partial context.} In IGMs, tokens are produced step by step, each prediction conditioned on previously generated content rather than modeling all tokens simultaneously. Consequently, the autoregressive transformer is tasked not only with determining the content of the tokens generated at the current step, but also with modeling their interdependence both within the simultaneously selected subset and across successive steps~\cite{li2025autoregressive}, so as to maintain semantic consistency under incomplete context. This more demanding objective may encourage the autoregressive transformer to learn signals that are stronger and sharper, which in turn may manifest as large-magnitude activations~\cite{shi2025meaningless}.

\textbf{(ii) Per-token continuous diffusion with gradients into the AR backbone.} Hybrid IGMs employ per-token DiffLoss to model $p(x\vert z)$ (Eq.~\ref{eq:diff_loss}), and the gradient of $z=f(\cdot)$ propagated from this loss optimizes the autoregressive transformer as well (Sec.~\ref{sec:MAR}), implying that the AR backbone is additionally supervised by the noise-regression objective. Furthermore, the same $z$ is reused for multiple sampled timesteps $t$~\cite{li2025autoregressive}, so a single forward pass optimizes AR representations to be effective across different noise strengths.
Taken together, these training signals can encourage the autoregressive transformer to encode conditioning vectors \(z\) that are more informative and more potent, potentially manifesting as surges in activations~\cite{shi2025meaningless}.

\textbf{(iii) Iterative composition and extended propagation path.} Both the autoregressive and diffusion proceed via iterative refinement. When composed, the overall process entails a longer sequence of updates~\cite{li2025autoregressive}, thereby extending the pathway along which errors may accumulate and propagate across steps~\cite{stine1987estimating,li2023error,1186529}. From a PTQ perspective~\cite{li2023q,liu2025error,so2023temporal}, this extended iterative trajectory can heighten sensitivity to quantization noise and render stable inference challenging. Therefore, compared to conventional ViTs, hybrid IGMs may be more susceptible to errors, presenting a more stringent challenge for PTQ.

To support the above analysis, Figure~\ref{figure_comparison} compares channel-wise maximum activations between hybrid IGMs and a conventional ViT in the same linear layer. 
Using the maximum activation observed in ViT as the threshold, hybrid IGMs exhibit sparse but much larger high-magnitude channels, indicating that outliers are significantly more pronounced in hybrid IGMs.

Based on this observation, we refer to these sparse high-magnitude activations as \textbf{Excessive Outliers}\textbf{ (EOs)} (Figure~\ref{figure2}(a)).
These outliers create a quantization trade-off, since quantization precision is determined by activation range (Eq.~\ref{quant_all}): either compress normal values into a narrow range to preserve outliers, or clip outliers to maintain normal precision. However, full-range alignment sacrifices normal values but impairs generation capability, while MSE-based \cite{chicco2021coefficient} truncation sacrifices outliers but irreversibly causes quality collapse (Figure~\ref{figure2}(b)(c)). 
These observations show that a single quantization scale fails to balance EOs and normal values. Our HCD (Sec.~\ref{HCD}) addresses this by hierarchically and adaptively decoupling them, preserving both distributions independently (Figure~\ref{figure2}(d)).

\subsection{Amplified Anomalies}\label{sec:AAs}

During quantized diffusion refinement in hybrid IGMs, we observe abnormal amplification of activations in the boundary layers, introducing severe difficulties for quantization.
To understand this phenomenon, we explicitly separate the quantization error in noise prediction from the denoising update.
Specifically, we denote the predicted noise by \(\hat{\varepsilon}_\theta(\hat{x}_t \vert t, z)\), and the quantization error by \(\Delta_{\varepsilon_\theta(\hat{x}_t \mid t, z)}\).
As per Eq.~(\ref{eq:denoising}), the sampling of \(\hat{x}_{t-1}\) in the quantized model can be expressed as:
\begin{align}
\hat{x}_{t-1} &= \frac{1}{\sqrt{\alpha_t}} \left( \hat{x_t} - \gamma_t \hat{\varepsilon}_\theta(\hat{x_t} \vert t, z) \right) + \sigma_t \delta  \\
&= \frac{1}{\sqrt{\alpha_t}} \Big( \hat{x_t} - \gamma_t \varepsilon_\theta(\hat{x_t} \vert t, z) - \gamma_t\Delta_{\varepsilon_\theta(\hat{x_t} \vert t, z)} \Big) + \sigma_t \delta.
\label{quantization_error}
\end{align}

In hybrid IGMs, activations at the boundary layers of the diffusion module are susceptible to abnormal amplification.
These boundary layers correspond to the input and output ends of the denoising update.
The final linear layer of the denoising MLP produces the predicted noise \(\hat{\varepsilon}_\theta(\hat{x}_t \vert t,z)\), and any quantization error in this prediction is injected into the reverse denoising step, where it is effectively scaled by \(\gamma_t\) when forming \(\hat{x}_{t-1}\). The resulting \(\hat{x}_{t-1}\) is then received by the input layer at the next timestep. In this sense, the final linear layer serves as the error-injection point, while the input layer directly receives the amplified state, making boundary-layer activations a direct indicator of error amplification during quantized diffusion refinement. Below, we analyze this mechanism from the following three aspects.

\textbf{(i) Stronger per-token guidance under incomplete context.}
In hybrid IGMs, the diffusion procedure models per-token distributions under the condition \(z\) produced by the autoregressive module. 
Since this condition is derived from an incomplete generation context, the denoising network needs to steer selected tokens from noisy states toward their semantic targets within limited sampling steps. 
To achieve this, the coefficient \(\gamma_t\) in Eq.~\ref{quantization_error} tends to be substantially larger at early timesteps~\cite{shi2025meaningless}, so even minor perturbations in \(\hat{\varepsilon}_\theta(\hat{x}_t \vert t,z)\) can induce an enlarged deviation in \(\hat{x}_{t-1}\).

\begin{figure}[t]
    \centering
  \includegraphics[width=0.48\textwidth]{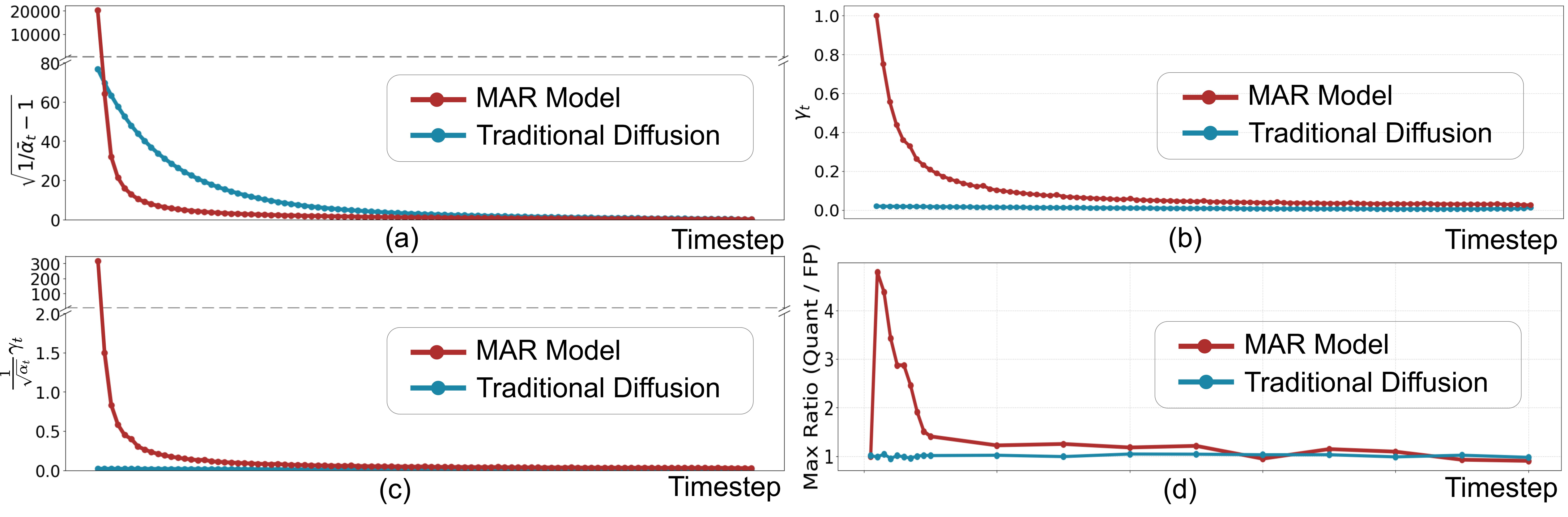}
  \vspace{-4mm}
  \caption{
  Comparison of denoising-related variables between hybrid IGM  and conventional diffusion models. 
  Subfigures (a), (b), and (c) show \(\sqrt{1-\bar{\alpha}_t}\), \(\gamma_t\), and \(\frac{1}{\sqrt{\alpha_t}}\gamma_t\), respectively. 
  Subfigure (d) reports the relative magnitude of boundary-layer inputs under quantization versus full precision.
  }
  \vspace{-5mm}
  \label{figure_comparison_AAs}
\end{figure}
\begin{figure}[t]
    \centering
  \includegraphics[width=0.48\textwidth]{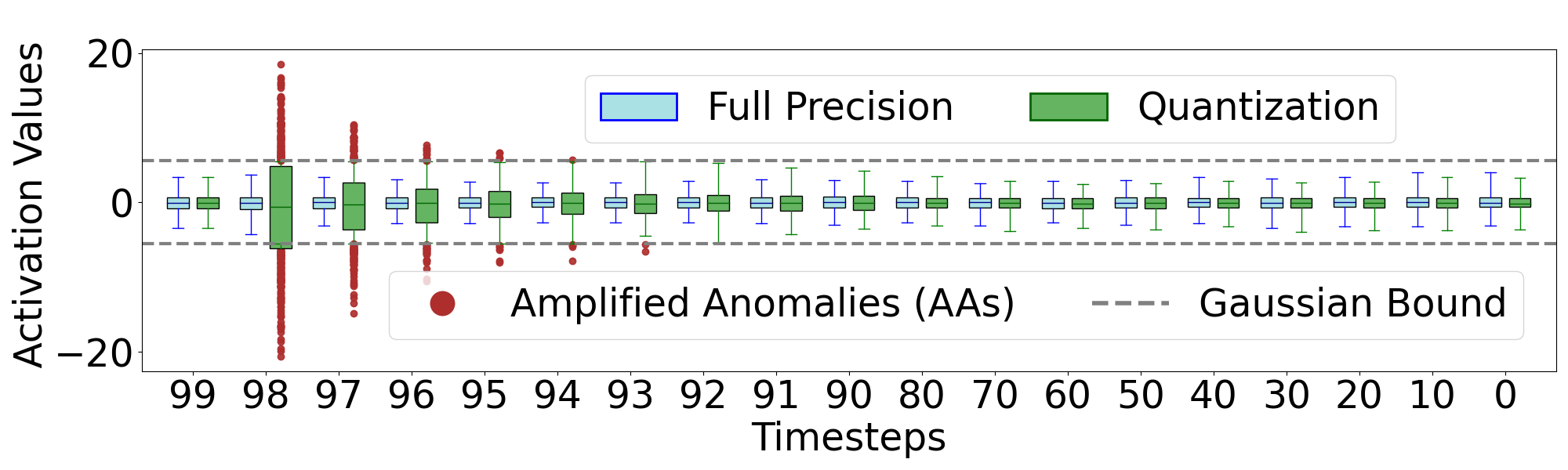}
  \vspace{-4mm}
  \caption{
  Boundary-layer activation distributions at selected denoising timesteps under full-precision and quantized inference. 
  The Gaussian Bound is estimated from the full-precision distribution (Sec.~\ref{sec:SR}); quantized activations beyond this bound are marked as \textcolor{myred}{\textbf{Amplified Anomalies (AAs)}}.
  }
  \vspace{-6mm}
  \label{figure4}
\end{figure}

\textbf{(ii) Amplification and propagation of quantization errors along hybrid iterations.}
The amplified denoising state does not remain isolated within a single diffusion step. 
Hybrid IGMs couple diffusion sampling with autoregressive context updates: the token produced by the diffusion procedure is incorporated into the subsequent context and then affects later conditional representations. 
Therefore, a local perturbation in one denoising step can be transformed into a biased condition for later token prediction~\cite{stine1987estimating,li2023error}. 
As autoregressive and diffusion iterations proceed jointly, such perturbations are repeatedly propagated and accumulated, making the quantized inference path increasingly deviate from the full-precision path~\cite{li2023q,liu2025error}.

\textbf{(iii) Shrinking quantization tolerance under amplified activations.}
Once minor quantization errors are amplified through the denoising update, the calibrated activation range leaves much less tolerance for deviations during inference.
Values that originally fall within a safe calibrated range may exceed the boundary after amplification, while tiny rounding errors~(Eq.~\ref{quant_all}) combined with stochasticity at each denoising step~(Eq.~\ref{eq:denoising}) make the effective activation range difficult to predict.
The resulting calibration-inference mismatch causes static quantization methods to handle amplified activations poorly.
Aggressive clipping disrupts the generation trajectory and substantially degrades perceptual quality.

To support the above analysis, Figure~\ref{figure_comparison_AAs} uses MAR as a representative hybrid IGM and compares its denoising behavior with that of conventional diffusion models. 
Figures~\ref{figure_comparison_AAs}(a)--(c) respectively report \(\sqrt{1-\bar{\alpha}_t}\), \(\gamma_t\), and \(\frac{1}{\sqrt{\alpha_t}}\gamma_t\) along the denoising process.
Here, \(\sqrt{1-\bar{\alpha}_t}\) controls the noise magnitude in \(x_t\) and appears in the clean-token reconstruction formula 
\(\hat{x}_0 = \frac{1}{\sqrt{\bar{\alpha}_t}}(\hat{x}_t-\sqrt{1-\bar{\alpha}_t}\cdot\varepsilon_\theta(\hat{x}_t\vert t,z))\).
Meanwhile, the larger \(\gamma_t\) and\(\frac{1}{\sqrt{\alpha_t}}\gamma_t\) at early timesteps indicate that the predicted noise, together with its quantization error, can exert a stronger influence on the denoising state \(\hat{x}_{t-1}\).
Figure~\ref{figure_comparison_AAs}(d) further shows that quantization enlarges the activations entering the boundary layers. 
To characterize these amplified activations, we use the full-precision activation distribution as a reference and define a Gaussian Bound (Sec.~\ref{sec:SR}) to represent the normal activation range (Figure~\ref{figure4}). 
Under quantized inference, activations that exceed this bound are regarded as out-of-range amplified activations, which we refer to as \textbf{Amplified Anomalies (AAs)}.

In summary, these anomalies suggest that hybrid IGMs are particularly vulnerable to quantization errors. 
Vanilla PTQ methods that rely on calibration-derived static clipping often lead to model failure.
To mitigate this, our Scaling Recalibration (Sec.~\ref{sec:SR}) dynamically detects anomalies and rescales them into a stable target range, especially at early timesteps, thus rescuing the risk of model collapse.

\begin{figure*}[t]
  \vspace*{-6mm}
  \centering
  \includegraphics[width=2\columnwidth]{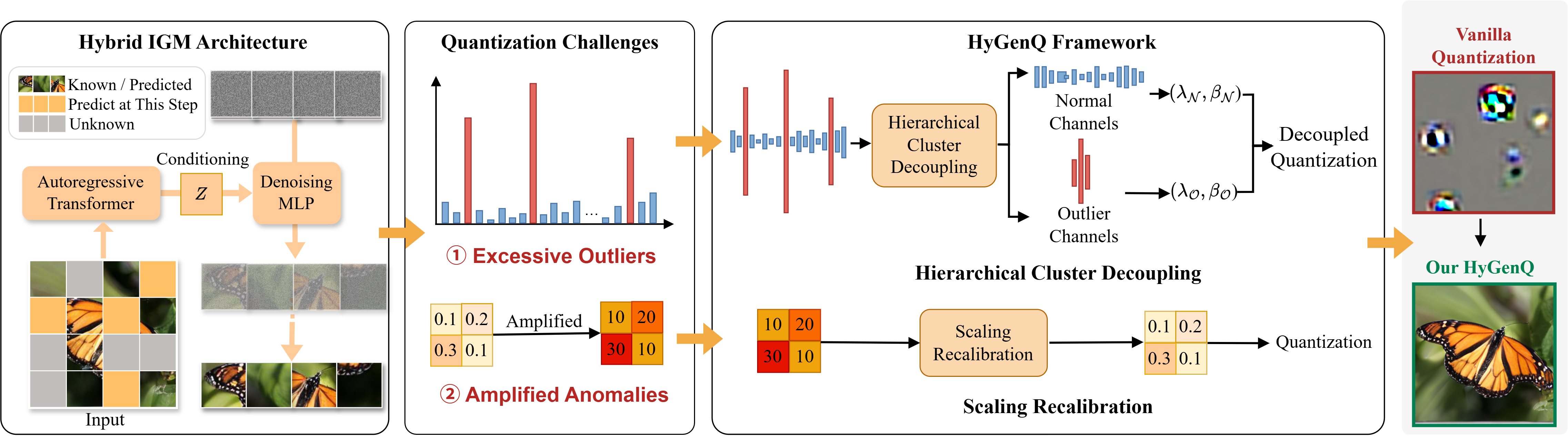}
  \vspace{-4mm}
  \caption{
  \textbf{Overview of the proposed HyGenQ framework.}
  Starting from the hybrid IGM architecture, we identify two quantization challenges: Excessive Outliers and Amplified Anomalies.
  To address them, HyGenQ introduces two complementary modules.
  Hierarchical Cluster Decoupling separates normal and outlier channels and assigns them independent quantization parameters, while Scaling Recalibration rescales amplified boundary-layer inputs before quantization.
  Together, they mitigate both EOs and AAs, enabling robust quantized inference for hybrid IGMs.
  }
  \vspace{-5mm}
  \label{figure_framework}
\end{figure*}

\vspace{-2mm}
\section{METHODOLOGY}
\label{method}
To address the quantization challenges of hybrid IGMs, we propose HyGenQ, a quantization framework consisting of two complementary modules: Hierarchical Cluster Decoupling (HCD) and Scaling Recalibration (SR). 
As illustrated in Figure~\ref{figure_framework}, HCD decouples normal and outlier channels to alleviate Excessive Outliers, while SR recalibrates amplified boundary-layer activations in the diffusion procedure to mitigate Amplified Anomalies.  
The details of the two modules are presented in Sec.~\ref{HCD} and Sec.~\ref{sec:SR}, respectively.
\vspace{-2mm}
\subsection{Hierarchical Cluster Decoupling}\label{HCD}
\begin{figure*}[t]
  \vspace*{-4mm}
  \centering
  \includegraphics[width=2\columnwidth]{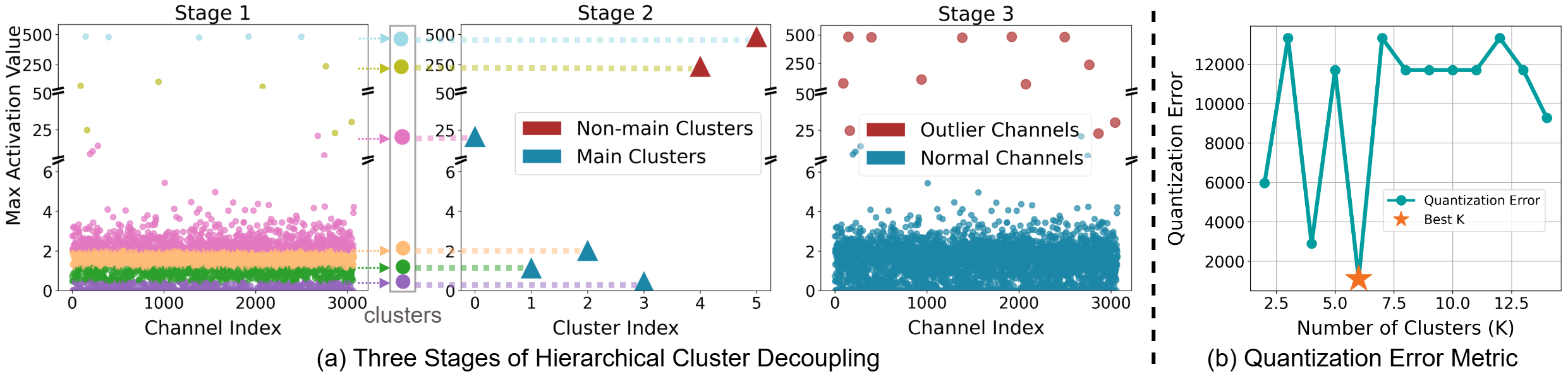}
  \vspace{-4mm}
  \caption{
\textbf{Illustration of Hierarchical Cluster Decoupling (HCD).}
(a) The module adopts a three-stage hierarchical clustering process to progressively identify and separate outlier channels from normal ones.
(b) The optimal clustering configuration is selected by minimizing a Quantization Error Metric based on the Sum of Squared Errors (SSE), yielding the cluster number \(k^*\) in Eq.~(\ref{optimal_k}).
}
  \vspace{-4mm}
  \label{figure_HCD}
\end{figure*}

As discussed in Section \ref{sec:EOs}, the coexistence of dominant normal values and outliers under a shared quantization scale leads to substantial errors. In hybrid IGMs, such errors accumulate across iterations, severely degrading generation quality.
Fortunately, we capitalize on the observation that these outliers consistently concentrate in a few fixed channels across iterations, which we define as \textit{outlier channels}, while others are treated as \textit{normal channels}.
Their stable spatial positions and consistent distribution patterns across iterations make it feasible to decouple the two types of channels for independent quantization. 
To this end, we propose Hierarchical Cluster Decoupling (HCD), which addresses this challenge through a three-stage procedure, as illustrated in Figure~\ref{figure_HCD}(a).

\textbf{Stage 1: Channel-Wise Feature Clustering.}
An FC2 linear layer in an autoregressive Transformer receives a token sequence as its input activation.
To facilitate processing, we reshape this activation along the channel dimension into a matrix \(\mathbf{X} \in \mathbb{R}^{n \times C}\), where \(C\) denotes the number of channels and each channel contains \(n\) elements for downstream processing.
In this form, each column \(\mathbf{X}_{i} \in \mathbb{R}^n\) represents the values of channel \(i\) across all positions.
Since quantization error is highly sensitive to the range of activations \cite{nagel2021white,xiao2023smoothquant,wu2025ptq4dit}, we extract a channel-wise feature vector \(\mathbf{c} = [c_1, c_2, \dots, c_C] \in \mathbb{R}^C\), where each component \(c_i\) is defined as the maximum absolute value in channel \(i\), computed as:
\begin{equation}
c_i = \max \left( \left| \mathbf{X}_{i} \right| \right), \quad i = 1, 2, \dots, C.
\label{channelwise_max}
\end{equation}

Then we form a compound feature set \(\mathcal{F} = \{(c_i, i)\}_{i=1}^C\)\, which is clustered into several groups based on feature similarity. To achieve this, we adopt \(k\)-means \cite{hartigan1979algorithm}, a classical algorithm widely used for efficient feature clustering \cite{yuan2023rptq,9248040,1377374,1413271}, as it minimizes intra-cluster feature variance. However, due to the highly imbalanced distribution of channel features, \(k\)-means is prone to suboptimal local minima \cite{ahmed2020k}. To mitigate this issue, we apply Genetic Algorithm \cite{krishna1999genetic,rahman2014hybrid} to enhance the robustness and global optimality of the clustering process.

Specifically, given a candidate cluster number $k$, we define a center proposal 
$\mathbf{Z}=\{\mu_j\}_{j=1}^k$ as a set of $k$ centers, each drawn from the channel-wise feature set $\{c_i\}_{i=1}^C$. 
We assess the quality of the proposal using the following objective as the fitness function~\cite{hartigan1979algorithm,kanungo2000analysis} for k-means:
\begin{equation}
\label{eq:ga_fitness_final}
\mathcal{J}(\mathbf{Z})=\sum_{i=1}^{C}\min_{1\le j\le k}\bigl|\,c_i-\mu_j\,\bigr|^2.
\end{equation}

Each proposal $\mathbf{Z}$ is treated as an individual in a population $\mathcal{P}^{(t)}$ of size $N_{pop}$. 
The population is evolved for $T$ generations through selection, crossover, and mutation, summarized as:
\begin{equation}
\label{eq:ga_update}
\mathcal{P}^{(t+1)}=
\mathcal{M}_{p_m}\!\Bigl(
\mathcal{X}_{p_c}\bigl(S(\mathcal{P}^{(t)};\,\mathcal{J})\bigr)
\Bigr).
\end{equation}
where $S$ denotes selection guided by fitness $\mathcal{J}$, 
$\mathcal{X}_{p_c}$ performs crossover with probability $p_c$, 
and $\mathcal{M}_{p_m}$ performs mutation with probability $p_m$\cite{krishna1999genetic}. 

After $T$ generations, the best individual $\mathbf{Z}^*$ is obtained as follows, and the final clustering is defined as 
\(\mathcal{C}=\{\mathcal{C}_j(\mathbf{Z}^*)\}_{j=1}^k\):
\vspace{-1mm}
\begin{equation}
\label{eq:best_individual}
\mathbf{Z}^*=\arg\min_{\mathbf{Z}\in\mathcal{P}^{(T)}}\mathcal{J}(\mathbf{Z}).
\end{equation}

Furthermore, to avoid the limitation of manually predefining the number of clusters \cite{hamerly2003learning,kodinariya2013review,pham2005selection}, we search over a practical range of candidate \(k\) values. For each \(k\), the corresponding clustering result is denoted as \(\mathcal{C}^{(k)} = \{ \mathcal{C}_1^{(k)}, \dots, \mathcal{C}_k^{(k)} \}\), where each \(\mathcal{C}_j^{(k)}\) represents a group of statistically similar channels.

\textbf{Stage 2: Cluster-Wise Aggregation and Partitioning.}
For each cluster $\mathcal{C}_j^{(k)}$, we compute its maximum value, denoted by $m_j^{(k)}$, and collect them into a vector $\mathbf{m}^{(k)} = [m_1^{(k)}, \dots, m_k^{(k)}]$. 
We then form a new compound set $\mathcal{G}^{(k)} = \{(m_j^{(k)}, j)\}_{j=1}^k$. 
Within the specified range of candidate $k$ values, each choice of $k$ leads to a second-stage clustering, where the set $\mathbf{m}^{(k)}$ is clustered into two partitions.
This clustering is formulated as:
\vspace{-2mm}

\begin{equation}
\label{eq:second_kmeans}
\mathcal{L}^{(k)}(\nu_1,\nu_2)
=\sum_{j=1}^{k}\min|m_j^{(k)}-\nu_{b_j}\bigr|^{2},\quad b_j \in \{1,2\}.
\end{equation}
where $b_j$ denotes the binary assignment of $m_j^{(k)}$ to one of the two groups with centers $\nu_1$ and $\nu_2$. 

Since each channel feature $c_i$ was represented by its maximum value in the previous stage, 
we designate as the main cluster the group whose mean maximum is smaller: 
\begin{equation}
\label{eq:choose_main_group_where1}
\begin{aligned}
\overline{m}_b = \frac{1}{|\{j:b_j=b\}|}\sum_{j:\,b_j=b} m_j^{(k)}, 
\quad 
{b\in\{1,2\}},
\end{aligned}
\end{equation}
\begin{equation}
\label{eq:choose_main_group_where2}
\begin{aligned}
b^\star &= \arg\min\, \overline{m}_b.
\end{aligned}
\end{equation}

Next, to enable subsequent quantization-error evaluation, we associate the main cluster with normal channels and the non-main cluster with outlier channels as follows:
\begin{equation}
\label{eq:map_back_channels}
\mathcal{N}^{(k)}=\!\!\bigcup_{j:\,b_j=b^\star}\!\!\mathcal{C}^{(k)}_j,
\quad
\mathcal{O}^{(k)}=\!\!\bigcup_{j:\,b_j\neq b^\star}\!\!\mathcal{C}^{(k)}_j.
\end{equation}

Since Stage 1 is repeated over \(k\), we establish a pseudo-quantization-based Quantization Error Metric to evaluate the quality of each clustering result.
Specifically, activation values from normal and outlier channels are pseudo-quantized with separate quantization parameters:
\begin{equation}
\widehat{\mathbf{X}}_i =
\begin{cases}
D\!\left(Q\!\left(\mathbf{X}_i;\lambda_{\mathcal{N}}^{(k)},\beta_{\mathcal{N}}^{(k)}\right)\right),
& i\in\mathcal{N}^{(k)},\\[1mm]
D\!\left(Q\!\left(\mathbf{X}_i;\lambda_{\mathcal{O}}^{(k)},\beta_{\mathcal{O}}^{(k)}\right)\right),
& i\in\mathcal{O}^{(k)}.
\end{cases}
\label{eq:hcd_pseudo_quant}
\end{equation}
where \((\lambda_{\mathcal{N}}^{(k)},\beta_{\mathcal{N}}^{(k)})\) and \((\lambda_{\mathcal{O}}^{(k)},\beta_{\mathcal{O}}^{(k)})\) are computed from the activation ranges of the normal and outlier channels, respectively.

With the above pseudo-quantized activations, we define a Quantization Error Metric using a weighted Sum of Squared Errors (SSE)~\cite{thinsungnoena2015clustering}, which emphasizes outlier-channel errors and avoids averaging out large errors from small outlier partitions:
\begin{equation}
\mathcal{L}^{(k)} =
\sum_{i\in\mathcal{O}^{(k)}} \left\| \mathbf{X}_i-\widehat{\mathbf{X}}_i \right\|_2^2
+
\omega
\sum_{i\in\mathcal{N}^{(k)}} \left\| \mathbf{X}_i-\widehat{\mathbf{X}}_i \right\|_2^2 .
\label{weighted_sse}
\end{equation}
\noindent where \(\omega\in[0,1]\) denotes the weighting factor for the normal-channel quantization error.
As shown in Figure~\ref{figure_HCD}(b), we determine the optimal cluster number \(k^*\) by minimizing \(\mathcal{L}^{(k)}\):
\begin{equation}
k^* = \arg\min \mathcal{L}^{(k)}.
\label{optimal_k}
\end{equation}

\textbf{Stage 3: Channel-Wise Outlier Assignment and Decoupling.}
Given the optimal \(\mathcal{O}^{(k^*)}\), we identify the outlier channels, while the remaining ones are treated as normal channels. 
Once identified, outlier channels are fixed per layer and reused during inference without recomputation.
We then perform decoupled quantization for the two partitions using the same bit-width but separate quantization parameters, which enables both parts to be quantized without mutual interference, thereby preserving the fine-grained precision of normal values while maintaining a sufficiently broad range for outliers.

\vspace{-2mm}
\subsection{Scaling Recalibration}\label{sec:SR}
\begin{figure*}[t]
  \centering
  \includegraphics[width=0.95\textwidth]{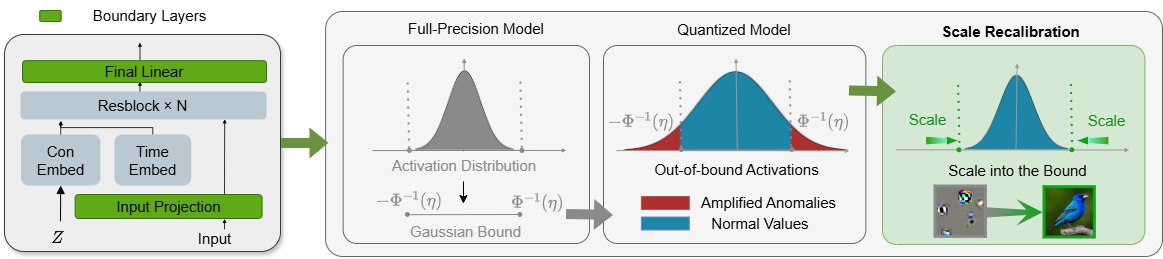}
  \vspace{-4mm}
  \caption{
  \textbf{Illustration of Scaling Recalibration (SR).}
  We apply Scaling Recalibration to the boundary-layer activations in the diffusion procedure of hybrid IGMs.
  Specifically, a Gaussian Bound is first defined from the activation distribution of the full-precision model, providing a stable reference range for quantization.
  During quantized inference, amplified quantization errors may cause boundary-layer activations to exceed the calibration range.
  The proposed strategy rescales these out-of-bound activations into the bound and applies this fixed range across timesteps, thereby stabilizing boundary-layer quantization. 
}
  \vspace{-5mm}
  \label{figure5}
\end{figure*}

During quantized inference, we apply Scaling Recalibration (SR) to the boundary layers in the diffusion procedure, as illustrated in Figure~\ref{figure5}.
These layers directly receive the intermediate denoising states, and thus small quantization errors can easily change their input activation ranges across timesteps.
Such range shifts create a mismatch between the calibration-time and inference-time activation distributions, making the quantization range unstable and difficult to predict.

In contrast, the boundary-layer input activations are sampled from a standard Gaussian distribution in the full-precision model (Sec.~\ref{sec:MAR}), which remains statistically stable across iterations and timesteps, as shown in Figure~\ref{figure4}.
Leveraging the inherent stability of the full-precision model, we define a high-confidence detection bound \(\Phi^{-1}(\eta)\), referred to as the \textit{Gaussian Bound}, where \(\Phi^{-1}(\cdot)\) denotes the inverse Cumulative Distribution Function (CDF) \cite{drew2000computing} of the standard normal distribution, and \(\eta\) is a chosen confidence level.

Since the Gaussian Bound defines the high-confidence range of full-precision activations, out-of-bound values in quantized models are treated as amplified anomalies and scaled accordingly.
Based on this, we utilize the \textit{Gaussian Bound} to identify anomalous channels.
Specifically, at each timestep $t$, if the maximum activation in channel $c$ of $\mathbf{X}_t$, denoted as $\max|\mathbf{X}^{c}_t|$, exceeds $\Phi^{-1}(\eta)$, we classify channel $c$ as anomalous. Formally, the set of such anomalous channels \(\mathcal{A}_t\) is defined as:
\begin{equation}
\mathcal{A}_t=\left\{ c \in \{1, \dots, C\} \;\middle|\; \max|\mathbf{X}^{c}_t| > \Phi^{-1}(\eta) \right\}.
\label{eq:scaling_channels}
\end{equation}

The channels in \(\mathcal{A}_t\) are rescaled by a shared factor \(s_{\mathcal{A}_t}\), which maps out-of-bound activations back into the \textit{Gaussian Bound}. 
The scaling process is defined as follows:
\vspace{-1mm}
\begin{equation}
\begin{split}
\widetilde{\mathbf{X}^{c}_t} = \frac{\mathbf{X}^{c}_t}{s_{\mathcal{A}_t}}, \quad
\text{where } s_{\mathcal{A}_t} = \max_{c \in \mathcal{A}_t} \left( \frac{\max|\mathbf{X}^{c}_t|}{\Phi^{-1}(\eta)} \right).
\label{eq:scaling_all}
\end{split}
\end{equation}

This formulation ensures activations remain within a stable range, thereby preventing harmful truncation during quantization. Inverse scaling is applied afterward to restore magnitudes.

While this scaling strategy addresses extreme anomalies, it is worth noting that for the majority of timesteps, activation values remain well within \(\Phi^{-1}(\eta)\). Nonetheless, distributional shifts between the calibration and inference data may still lead to moderate activation range expansion, which, under standard quantization settings, could result in suboptimal clipping (Eq. \ref{quant_all}).
To ensure robustness without relying on dynamic scaling, SR adopts a fixed-range quantization strategy based on \(\Phi^{-1}(\eta)\). In this mode, rather than using sample-dependent extrema, the quantization range (Eq. \ref{quant_all}) is defined as \([-\Phi^{-1}(\eta), \Phi^{-1}(\eta)]\), from which \(\lambda\) and \(\beta\) are derived as follows:
\begin{equation}
\lambda = \frac{\Phi^{-1}(\eta)-\left(-\Phi^{-1}(\eta)\right)}{2^b - 1}, 
\quad 
\beta = - \left\lfloor \frac{-\Phi^{-1}(\eta)}{\lambda} \right\rceil.
\label{quant_sr}
\end{equation}

\section{Experiments}
\label{experiments}
\subsection{Experimental Settings}\label{sec:Experimental Settings}
\textbf{Implementation Protocol.}  
We generate 50,000 images for each method on 256×256 ImageNet~\cite{deng2009imagenet}, following the official configurations of all models and using identical iterative generation settings across methods.
All experiments are implemented using the PyTorch framework~\cite{paszke2019pytorch} and executed on a single NVIDIA RTX 4090 GPU.  
To assess generation quality, we adopt two standard evaluation metrics: Fréchet Inception Distance (FID)~\cite{heusel2017gans} and Inception Score (IS)~\cite{salimans2016improved}, both computed using the official ADM evaluation toolkit~\cite{dhariwal2021diffusion}.

\begin{table*}[t]
\caption{Performance comparison with 64 iterations and 100 timesteps, where \spc indicates that the boundary layers in the diffusion procedure remain in full precision, while \pct denotes that all linear layers are quantized.
}
\centering
\vspace*{-3mm}
\label{table1}
\tiny
\begingroup
\setlength{\tabcolsep}{5pt}
\renewcommand{\arraystretch}{1.08}
\resizebox{0.95\textwidth}{!}{
\small
\begin{tabular}{lccccccccc}
\toprule
Method & Model & Precision (W/A) & Iteration & Timestep & Size
& \spc FID $\downarrow$ & \spc IS $\uparrow$
& \pct FID $\downarrow$ & \pct IS $\uparrow$ \\
\midrule
\textcolor{gray}{Full-Precision} & \textcolor{gray}{MAR-B} & \textcolor{gray}{32/32} & \textcolor{gray}{64} & \textcolor{gray}{100} & \textcolor{gray}{208M}
& \textcolor{gray}{2.30} & \textcolor{gray}{279.44}
& \textcolor{gray}{2.30} & \textcolor{gray}{279.44} \\
\midrule
UniformQuant & MAR-B & 8/8 & 64 & 100 & 52M & 237.59 & 2.51 & -- & -- \\
RepQ-ViT \cite{li2023repq} & MAR-B & 8/8 & 64 & 100 & 52M & 298.71 & 1.60 & -- & -- \\
SmoothQuant \cite{xiao2023smoothquant} & MAR-B & 8/8 & 64 & 100 & 52M & 10.31 & 144.51 & 253.28 & 1.97 \\
PTQ4DM \cite{shang2023ptq4dm} & MAR-B & 8/8 & 64 & 100 & 52M & 10.34 & 144.22 & 252.41 & 1.98 \\
TFMQ-DM \cite{huang2024tfmq} & MAR-B & 8/8 & 64 & 100 & 52M & 9.22 & 150.36 & 256.17 & 1.90 \\
TaQ-DiT \cite{11342323} & MAR-B & 8/8 & 64 & 100 & 52M & 5.38 & 198.05 & 143.85 & 2.52 \\
OCS \cite{zhao2019improving} & MAR-B & 8/8 & 64 & 100 & 52M & 5.89 & 185.53 & 251.75 & 1.98 \\
\textbf{HyGenQ (Ours)} & MAR-B & 8/8 & 64 & 100 & 52M & \textbf{3.50} & \textbf{247.33} & \textbf{3.61} & \textbf{246.00} \\
\midrule
\textcolor{gray}{Full-Precision} & \textcolor{gray}{MAR-L} & \textcolor{gray}{32/32} & \textcolor{gray}{64} & \textcolor{gray}{100} & \textcolor{gray}{479M}
& \textcolor{gray}{1.80} & \textcolor{gray}{294.10}
& \textcolor{gray}{1.80} & \textcolor{gray}{294.10} \\
\midrule
UniformQuant & MAR-L & 8/8 & 64 & 100 & 120M & 396.28 & 1.60 & -- & -- \\
RepQ-ViT \cite{li2023repq} & MAR-L & 8/8 & 64 & 100 & 120M & 437.30 & 1.60 & -- & -- \\
SmoothQuant \cite{xiao2023smoothquant} & MAR-L & 8/8 & 64 & 100 & 120M & 6.07 & 193.11 & 247.86 & 1.98 \\
PTQ4DM \cite{shang2023ptq4dm} & MAR-L & 8/8 & 64 & 100 & 120M & 6.13 & 193.90 & 248.26 & 1.97 \\
TFMQ-DM \cite{huang2024tfmq} & MAR-L & 8/8 & 64 & 100 & 120M & 6.28 & 187.17 & 250.56 & 1.99 \\
TaQ-DiT \cite{11342323} & MAR-L & 8/8 & 64 & 100 & 120M & 4.99 & 208.68 & 139.90 & 2.68 \\
OCS \cite{zhao2019improving} & MAR-L & 8/8 & 64 & 100 & 120M & 3.69 & 222.45 & 245.58 & 2.06 \\
\textbf{HyGenQ (Ours)} & MAR-L & 8/8 & 64 & 100 & 120M & \textbf{2.91} & \textbf{259.59} & \textbf{3.14} & \textbf{255.35} \\
\midrule
\textcolor{gray}{Full-Precision} & \textcolor{gray}{MAR-H} & \textcolor{gray}{32/32} & \textcolor{gray}{64} & \textcolor{gray}{100} & \textcolor{gray}{943M}
& \textcolor{gray}{1.58} & \textcolor{gray}{299.72}
& \textcolor{gray}{1.58} & \textcolor{gray}{299.72} \\
\midrule
UniformQuant & MAR-H & 8/8 & 64 & 100 & 236M & 564.27 & 1.00 & -- & -- \\
RepQ-ViT \cite{li2023repq} & MAR-H & 8/8 & 64 & 100 & 236M & 564.27 & 1.00 & -- & -- \\
SmoothQuant \cite{xiao2023smoothquant} & MAR-H & 8/8 & 64 & 100 & 236M & 9.28 & 196.03 & 267.68 & 1.84 \\
PTQ4DM \cite{shang2023ptq4dm} & MAR-H & 8/8 & 64 & 100 & 236M & 9.07 & 200.92 & 266.10 & 1.84 \\
TFMQ-DM \cite{huang2024tfmq} & MAR-H & 8/8 & 64 & 100 & 236M & 6.22 & 219.58 & 257.85 & 1.92 \\
TaQ-DiT \cite{11342323} & MAR-H & 8/8 & 64 & 100 & 236M & 4.96 & 226.37 & 156.23 & 2.34 \\
OCS \cite{zhao2019improving} & MAR-H & 8/8 & 64 & 100 & 236M & 4.25 & 240.68 & 261.37 & 1.89 \\
\textbf{HyGenQ (Ours)} & MAR-H & 8/8 & 64 & 100 & 236M & \textbf{3.77} & \textbf{257.31} & \textbf{6.89} & \textbf{210.95} \\
\bottomrule
\end{tabular}
}
\endgroup
\vspace{-5mm}
\end{table*}

\textbf{Baseline Configuration.}
To evaluate the effectiveness of HyGenQ, we compare it against several representative PTQ methods for ViTs and diffusion models.
\textbf{RepQ-ViT}~\cite{li2023repq} utilizes scale reparameterization to address distribution mismatches in LayerNorm and Softmax activations.
\textbf{SmoothQuant}~\cite{xiao2023smoothquant} shifts the quantization burden from activations to weights, thereby mitigating activation-induced quantization difficulty.
\textbf{PTQ4DM}~\cite{shang2023ptq4dm} considers activation variance across denoising steps with reconstruction-based quantization.
\textbf{TFMQ-DM}~\cite{huang2024tfmq} preserves timestep-dependent temporal information by introducing Temporal Information Blocks.
\textbf{TaQ-DiT}~\cite{11342323} improves DiT quantization by mitigating the irregular and outlier-prone distributions of sensitive Post-GELU activations.
\textbf{OCS}~\cite{zhao2019improving} mitigates outlier-induced quantization error by splitting extreme-value channels.
Furthermore, we use \textbf{UniformQuant} as the basic quantization baseline and perform outlier truncation by minimizing the Mean Squared Error (MSE)~\cite{chicco2021coefficient}.
All baselines are reimplemented and adapted to the MAR architecture while preserving their original methodological designs.

\textbf{Calibration Set.}
We construct the calibration set by randomly selecting 32 samples for 256×256 ImageNet~\cite{deng2009imagenet} evaluation. During calibration, we run the full hybrid generation process on these samples and collect activation statistics across all autoregressive iterations and diffusion timesteps. We use the same calibration set and sampling protocol for all baseline methods and HyGenQ to ensure fair comparisons.

\textbf{Quantization Configuration.}
All experiments are conducted under deterministic settings to ensure fair and reproducible comparisons. Unless otherwise specified, a uniform quantizer is applied to all weights and activations. 
To control activation shifts across autoregressive iterations and diffusion timesteps~\cite{sun2024tmpq}, we assign separate quantization parameters to each iteration and timestep, enabling a focused evaluation of excessive outliers and amplified anomalies.
This quantization strategy is consistently adopted across all baseline methods and HyGenQ to ensure fair comparisons.

\textbf{HyGenQ Configuration.}  
Our HyGenQ framework is composed of two components: Hierarchical Cluster Decoupling (HCD) and Scaling Recalibration (SR).  
For HCD, the default range of candidate cluster numbers is set to $k\in[2,20]$, and the weighting factor is fixed to $\omega=0.001$. The genetic algorithm follows the DEAP implementation \cite{fortin2012deap}, with population size $P=50$, number of generations $T=20$, crossover probability $p_c=0.5$, and mutation probability $p_m=0.2$.  
For MAR-H, SR is additionally enabled on boundary layers during the first 50\% of the denoising timesteps in each iteration to mitigate activation overflow and undesirable clipping, while avoiding extra computation in later timesteps.

\begin{figure*}[t]
\vspace{-4mm}
  \centering
  \begin{minipage}[t]{0.65\textwidth}
    \centering
    \captionof{table}{Cross-architecture generalization of our HyGenQ on ImageNet \(256\times256\)~\cite{deng2009imagenet}.}
    \vspace{-2mm}
    \label{table:generalization}
    \small
    \setlength{\tabcolsep}{5pt}
    \renewcommand{\arraystretch}{1.15}
    \begin{tabular}{lccccccc}
      \toprule
      Model & Size & Method & IS $\uparrow$ & FID $\downarrow$ & sFID $\downarrow$ & Precision $\uparrow$ & Recall $\uparrow$ \\
      \midrule
      \multirow{9}{*}{\shortstack[c]{LlamaGen\\ \cite{sun2024autoregressive}}} & \multirow{3}{*}{B} & \textcolor{gray}{Full-Precision } & \textcolor{gray}{151.948} & \textcolor{gray}{10.154} & \textcolor{gray}{25.625} & \textcolor{gray}{0.818} & \textcolor{gray}{0.571} \\
      & & PTQ4DM\cite{shang2023ptq4dm} & 140.898 & 10.794 & \textbf{26.704} & 0.787 & 0.590 \\
      & & \textbf{HyGenQ (Ours)} & \textbf{145.026} & \textbf{10.450} & 26.951 & \textbf{0.801} & \textbf{0.589} \\
      \cmidrule(lr){2-8}
      & \multirow{3}{*}{L} & \textcolor{gray}{Full-Precision } & \textcolor{gray}{250.294} & \textcolor{gray}{6.911} & \textcolor{gray}{23.942} & \textcolor{gray}{0.839} & \textcolor{gray}{0.635} \\
      & & PTQ4DM\cite{shang2023ptq4dm} & 218.392 & \textbf{7.107} & 24.652 & 0.800 & 0.670 \\
      & & \textbf{HyGenQ (Ours)} & \textbf{221.157} & 7.150 & \textbf{24.283} & \textbf{0.800} & \textbf{0.670} \\
      \cmidrule(lr){2-8}
      & \multirow{3}{*}{XL} & \textcolor{gray}{Full-Precision } & \textcolor{gray}{282.549} & \textcolor{gray}{6.713} & \textcolor{gray}{23.407} & \textcolor{gray}{0.827} & \textcolor{gray}{0.655} \\
      & & PTQ4DM\cite{shang2023ptq4dm} & 242.487 & 7.001 & 25.236 & 0.770 & 0.694 \\
      & & \textbf{HyGenQ (Ours)} & \textbf{252.308} & \textbf{6.598} & \textbf{24.070} & \textbf{0.793} & \textbf{0.678} \\
      \midrule
      \multirow{3}{*}{\shortstack[c]{LDM-4\\{~\cite{rombach2022high}}}} & \multirow{3}{*}{-} & \textcolor{gray}{Full-Precision } & \textcolor{gray}{358.030} & \textcolor{gray}{17.874} & \textcolor{gray}{36.598} & \textcolor{gray}{0.941} & \textcolor{gray}{0.544} \\
      & & PTQ4DM\cite{shang2023ptq4dm} & 357.602 & \textbf{17.271} & 35.672 & 0.939 & 0.548 \\
      & & \textbf{HyGenQ (Ours)} & \textbf{360.513} & \textbf{17.457} & \textbf{35.569} & \textbf{0.941} & \textbf{0.548} \\
      \bottomrule
    \end{tabular}
  \end{minipage}
  \hfill
  \begin{minipage}[t]{0.295\textwidth}
    \centering
    \vspace*{0.8\baselineskip} 
    \includegraphics[width=0.9\linewidth]{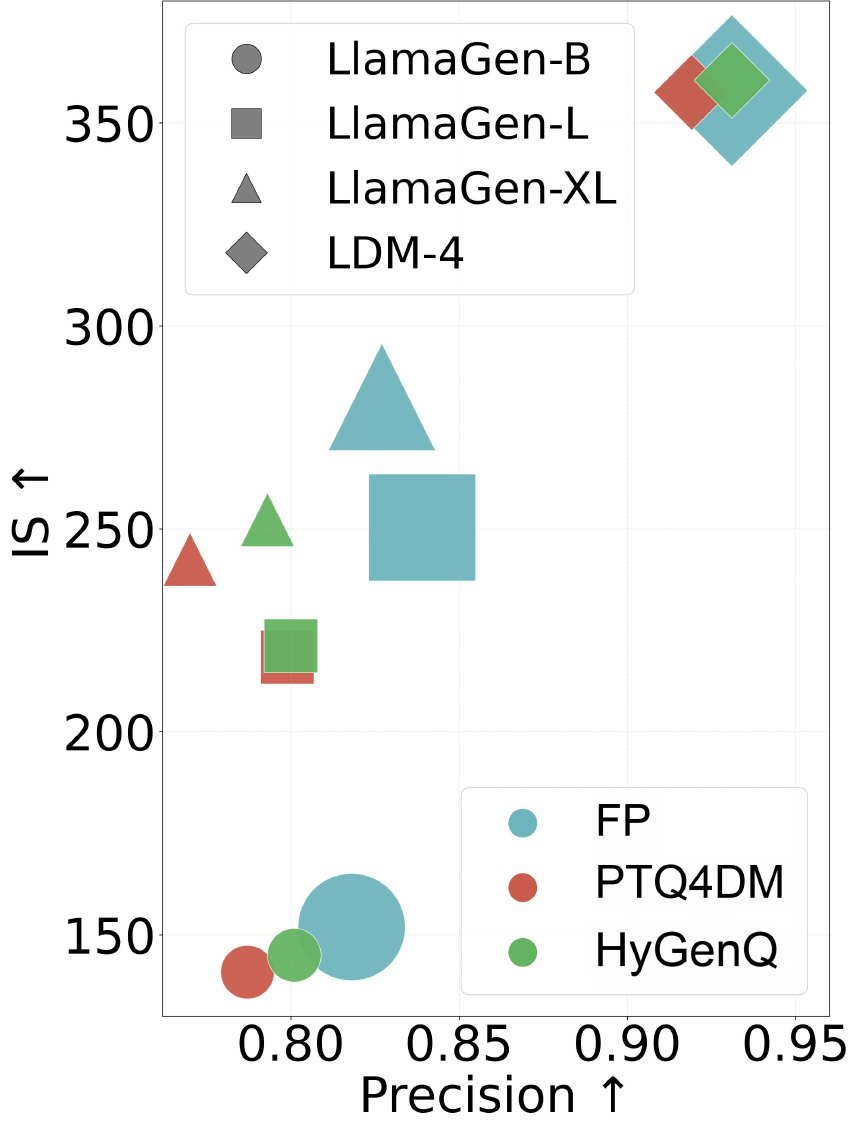}
    \vspace{-4mm}
    \captionof{figure}{Quantization performance on different models. Circle size indicates model size.}
    \label{fig:generalization}
  \end{minipage}
\vspace{-4mm}
\end{figure*}

\begin{table}[t]
\centering
\caption{Performance comparison with 32 iterations and 100 timesteps. Marker \spc indicates that the boundary layers in the diffusion procedure remain in full precision, while marker \pct denotes that all linear layers are quantized.
}
\vspace{-2mm}
\label{table2}
\resizebox{1\columnwidth}{!}{%
\small
\begin{tabular}{lcccc}
\toprule
Method & Model & Iteration & FID $\downarrow$ & IS $\uparrow$ \\
\midrule
\textcolor{gray}{Full-Precision} & \textcolor{gray}{MAR-B} & \textcolor{gray}{32} & \textcolor{gray}{2.46} & \textcolor{gray}{274.23} \\
\midrule
\spc SmoothQuant \cite{xiao2023smoothquant} & MAR-B & 32 & 11.93 & 139.61 \\
\spc PTQ4DM \cite{shang2023ptq4dm} & MAR-B & 32 & 11.88 & 139.66 \\
\spc \textbf{HyGenQ (Ours)} & MAR-B & 32 & \textbf{4.36} & \textbf{240.05} \\
\midrule
\pct SmoothQuant \cite{xiao2023smoothquant} & MAR-B & 32 & 202.44 & 2.99 \\
\pct PTQ4DM \cite{shang2023ptq4dm} & MAR-B & 32 & 207.28 & 2.88 \\
\pct \textbf{HyGenQ (Ours)} & MAR-B & 32 & \textbf{4.44} & \textbf{236.79} \\
\midrule
\textcolor{gray}{Full-Precision} & \textcolor{gray}{MAR-L} & \textcolor{gray}{32} & \textcolor{gray}{2.08} & \textcolor{gray}{282.15} \\
\midrule
\spc SmoothQuant \cite{xiao2023smoothquant} & MAR-L & 32 & 7.32 & 187.43 \\
\spc PTQ4DM \cite{shang2023ptq4dm} & MAR-L & 32 & 7.42 & 185.23 \\
\spc \textbf{HyGenQ (Ours)} & MAR-L & 32 & \textbf{3.52} & \textbf{250.12} \\
\midrule
\pct SmoothQuant \cite{xiao2023smoothquant} & MAR-L & 32 & 190.75 & 3.45 \\
\pct PTQ4DM \cite{shang2023ptq4dm} & MAR-L & 32 & 190.61 & 3.46 \\
\pct \textbf{HyGenQ (Ours)} & MAR-L & 32 & \textbf{3.80} & \textbf{243.83} \\
\midrule
\textcolor{gray}{Full-Precision} & \textcolor{gray}{MAR-H} & \textcolor{gray}{32} & \textcolor{gray}{1.95} & \textcolor{gray}{283.24} \\
\midrule
\spc SmoothQuant \cite{xiao2023smoothquant} & MAR-H & 32 & 11.21 & 183.89 \\
\spc PTQ4DM \cite{shang2023ptq4dm} & MAR-H & 32 & 11.06 & 186.27 \\
\spc \textbf{HyGenQ (Ours)} & MAR-H & 32 & \textbf{5.15} & \textbf{239.69} \\
\midrule
\pct SmoothQuant \cite{xiao2023smoothquant} & MAR-H & 32 & 208.14 & 2.79 \\
\pct PTQ4DM \cite{shang2023ptq4dm} & MAR-H & 32 & 208.38 & 2.80 \\
\pct \textbf{HyGenQ (Ours)} & MAR-H & 32 & \textbf{15.64} & \textbf{151.89} \\
\bottomrule
\end{tabular}
}
\vspace{-6mm}
\end{table}

\subsection{Quantization Performance of Hybrid IGMs}
\label{sec:Quantization Performance}
\begin{figure}[h]
  \centering
  \vspace*{-4mm}
  \includegraphics[width=0.48\textwidth]{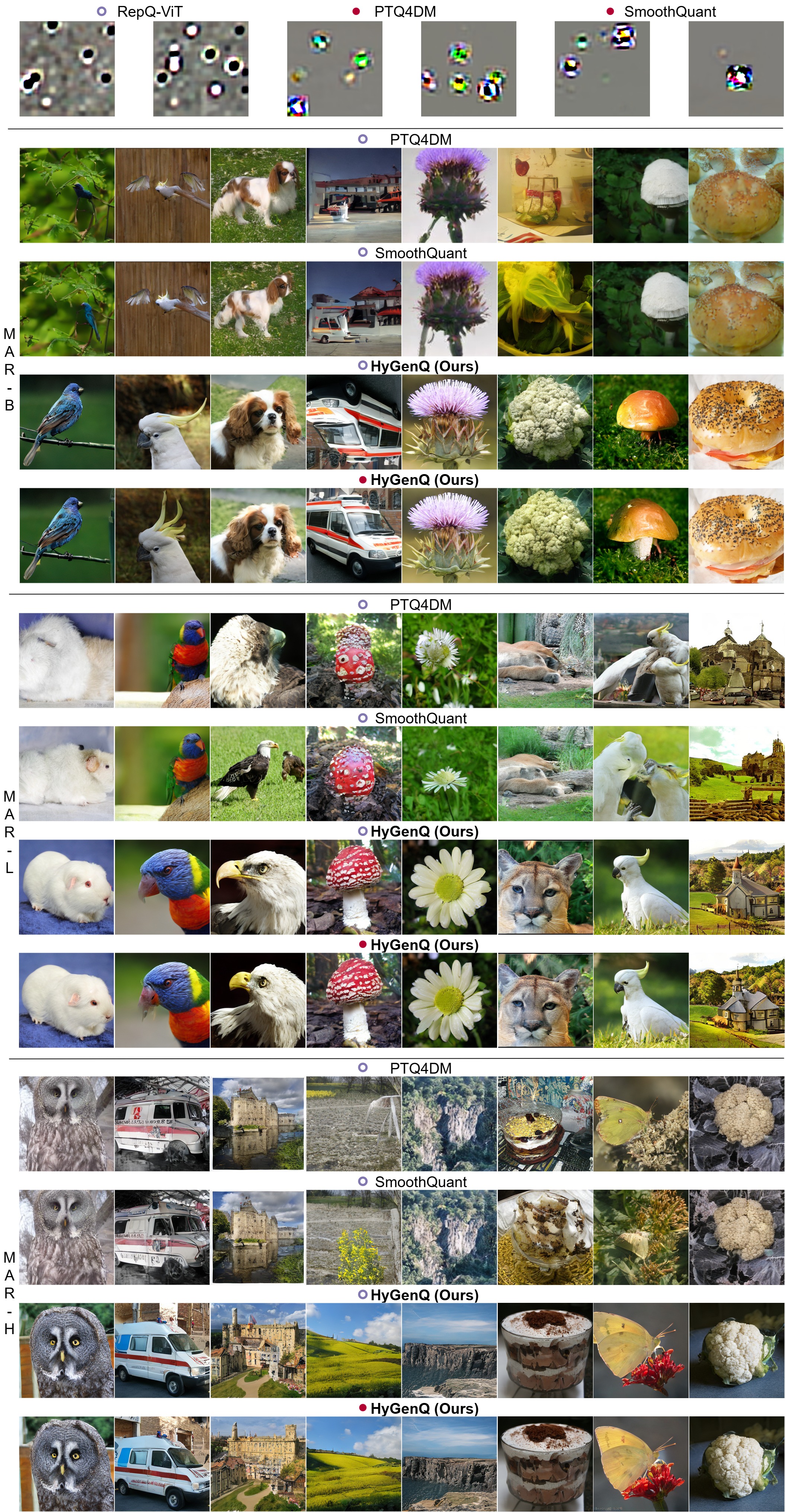}
  \vspace{-2mm}
  \caption{Visualization results with 64 iterations and 100 timesteps in MAR models. Marker \spc indicates that the boundary layers in diffusion remain in full precision, while \pct denotes that all linear layers are quantized. HyGenQ outperforms all baselines in both settings. Please zoom in for details.
  }
  \vspace{-3mm}
  \label{figure6}
\end{figure}

Given that hybrid IGMs exhibit the most severe error accumulation due to their tightly coupled multi-step inference, we use MAR model as a representative hybrid instantiation for our primary evaluation.
We evaluate all methods under two settings: 64 autoregressive iterations with a 100-step DDPM solver~\cite{ho2020denoising}, reported quantitatively in Table~\ref{table1} and visually in Figure~\ref{figure6}; and 32 iterations with 100 steps to assess robustness, reported in Table~\ref{table2}.
Both tables report results under two quantization settings: \spc keeps the boundary layers in the diffusion procedure in full precision, which weakens degradation caused by amplified anomalies and better isolates how different methods handle excessive outliers, while \pct quantizes all linear layers.

Under the \spc setting, UniformQuant and RepQ-ViT~\cite{li2023repq} collapse across model scales, with MAR-H reaching FID 564.27 and IS 1.00.
This shows that MSE-based clipping or normalization-oriented reparameterization cannot preserve the informative outliers required by hybrid generation.
Other baselines alleviate the degradation to some extent, but remain limited.
SmoothQuant~\cite{xiao2023smoothquant} shifts activation difficulty to weights, which can introduce additional weight-side outliers in MAR.
PTQ4DM~\cite{shang2023ptq4dm} retains more high-magnitude activation information than direct clipping, but still leaves the large magnitude gap between outliers and normal activations unresolved.
TFMQ-DM~\cite{huang2024tfmq} preserves timestep-dependent temporal information through Temporal Information Blocks, while this is not the dominant bottleneck in hybrid IGMs.
TaQ-DiT~\cite{11342323} mitigates irregular Post-GELU outliers, but its conservative strategy still leaves high-magnitude activations that dominate the quantization scale and suppress normal activations.
OCS~\cite{zhao2019improving} further reduces outlier effects by splitting extreme-value channels, but its fixed outlier selection is less adaptive to the varying activation patterns in hybrid generation.
In contrast, HyGenQ achieves the best performance across all models under the \spc setting, reaching FID 2.91 and IS 259.59 on MAR-L.
This indicates that HyGenQ better preserves both informative excessive outliers and normal activation values, thereby retaining critical generative information.

\begin{figure}[t]
    \centering
    \includegraphics[width=0.48\textwidth]{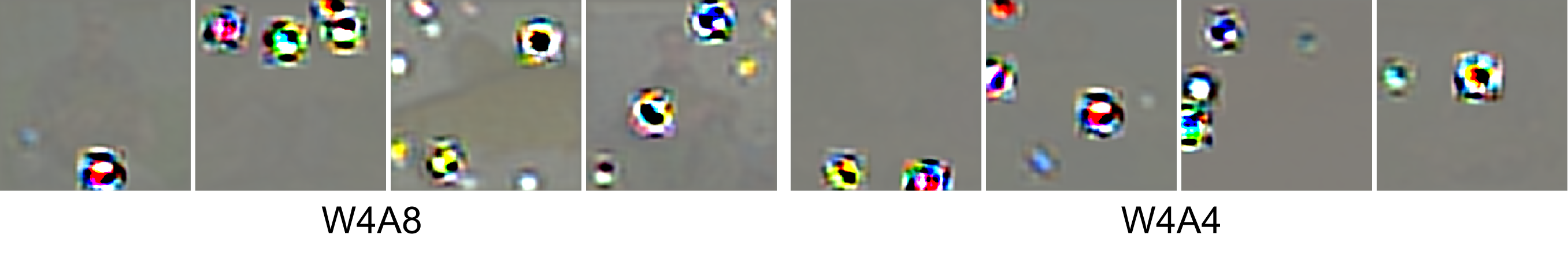}
    \vspace{-4mm}
    \caption{
    Visualization results under lower-bit quantization. W4A8 and W4A4 cause severe generation collapse for most methods.
    }
    \vspace{-5mm}
    \label{lower}
\end{figure}

We further evaluate the more stringent \pct setting, where all linear layers are quantized.
This setting exposes AA-induced error accumulation because anomalies can no longer be suppressed by keeping boundary layers in full precision.
Among the remaining baselines, SmoothQuant, PTQ4DM, TFMQ-DM, and OCS all collapse on MAR-L, with FID values around 250 and IS close to 2, while TaQ-DiT also degrades to FID 139.90 and IS 2.68.
These results indicate that handling activation outliers alone is insufficient when all linear layers are quantized, since anomalies can be repeatedly amplified throughout the multi-step inference process.
HyGenQ remains stable under this challenging setting across all MAR scales.
Notably, it achieves FID 3.14 and IS 255.35 on MAR-L, and even on the deeper MAR-H, it maintains a competitive FID of 6.89 and IS of 210.95.
This indicates that HyGenQ effectively suppresses the anomaly propagation and limits accumulated degradation.

\vspace{-4mm}
\subsection{Robustness and Generalization Across IGM Families}\label{sec:robustness_generalization}

To evaluate the robustness and generalization of HyGenQ across diverse IGM paradigms, we conduct experiments on representative autoregressive and diffusion models in addition to hybrid IGMs.
All experiments use ImageNet \(256\times256\)~\cite{deng2009imagenet}, generating 8{,}000 samples per model.
Notably, metrics computed with different sample counts should not be directly compared across tables.
Following a controlled protocol, we keep the calibration data, bit-width configuration, and sampling settings consistent across methods.
Unless otherwise specified, all baselines are built upon PTQ4DM~\cite{shang2023ptq4dm}.

Specifically, \textbf{LlamaGen}~\cite{sun2024autoregressive} represents the autoregressive paradigm, where token-by-token generation can exhibit heavy-tailed activations and sensitivity to range selection under quantization.
\textbf{LDM-4}~\cite{rombach2022high} represents the diffusion paradigm, whose iterative denoising procedure can accumulate quantization perturbations across timesteps.
As shown in Table~\ref{table:generalization} and Figure~\ref{fig:generalization}, HyGenQ consistently improves over PTQ4DM across these settings.
In several cases, HyGenQ outperforms the full-precision model on certain metrics, suggesting that HyGenQ can generalize beyond hybrid IGMs.

\begin{table}[t]
\centering
\captionsetup{width=1\linewidth}
\caption{Additional comparison with outlier-suppression quantization methods on ImageNet \(256\times256\) with 8,000 generated samples.}
\label{table:hadanorm_quarot}
\vspace{-2mm}
\begingroup
\setlength{\tabcolsep}{3pt}
\renewcommand{\arraystretch}{1.05}
\resizebox{0.95\linewidth}{!}{
\begin{tabular}{lccccc}
\toprule
Model & Method & \spc FID $\downarrow$ & \spc IS $\uparrow$ & \pct FID $\downarrow$ & \pct IS $\uparrow$ \\
\midrule
MAR-B & QuaRot~\cite{ashkboos2024quarot} & 7.28 & 172.28 & 221.84 & 1.95 \\
MAR-B & HadaNorm~\cite{federici2025hadanorm} & \textbf{6.95} & 173.74 & 214.37 & 2.11 \\
MAR-B & \textbf{HyGenQ (Ours)} & 7.14 & \textbf{176.70} & \textbf{7.45} & \textbf{162.28} \\
\midrule
MAR-L & QuaRot~\cite{ashkboos2024quarot} & 5.96 & 187.90 & 216.92 & 2.08 \\
MAR-L & HadaNorm~\cite{federici2025hadanorm} & \textbf{5.95} & 187.68 & 208.56 & 2.24 \\
MAR-L & \textbf{HyGenQ (Ours)} & 6.06 & \textbf{193.94} & \textbf{6.54} & \textbf{190.17} \\
\midrule
MAR-H & QuaRot~\cite{ashkboos2024quarot} & \textbf{6.61} & 188.54 & 233.75 & 1.79 \\
MAR-H & HadaNorm~\cite{federici2025hadanorm} & 6.67 & \textbf{189.79} & 226.41 & 1.88 \\
MAR-H & \textbf{HyGenQ (Ours)} & 7.19 & 188.32 & \textbf{13.03} & \textbf{154.39} \\
\bottomrule
\end{tabular}
}
\endgroup
\vspace{-4mm}
\end{table}

\vspace{-3mm}
\subsection{Extending Outlier Suppression to Hybrid IGMs}

To further verify the effectiveness of HyGenQ on hybrid IGMs, we transfer two representative outlier-suppression quantization methods, QuaRot~\cite{ashkboos2024quarot} and HadaNorm~\cite{federici2025hadanorm}, to MAR.
As shown in Table~\ref{table:hadanorm_quarot}, both methods achieve competitive results under the \spc setting, indicating that redistributing activation outliers can partially benefit hybrid IGM quantization.
However, their performance remains unstable across model scales, since they mainly alleviate outlier effects rather than fully resolving the large gap between excessive outliers and normal activation values.
Under the more stringent \pct setting, both methods suffer severe collapse, showing that outlier suppression alone cannot handle amplified anomalies and accumulated error propagation when all linear layers are quantized.
By contrast, HyGenQ remains much more stable, further demonstrating its effectiveness for robust hybrid IGM quantization.

\subsection{Ablation Study}\label{sec:Ablation Study}

\begin{table}[t]
\centering
\caption{Ablation study on ImageNet 256×256 with 64 iterations and 100 timesteps in MAR-B.}
\label{table ablation 1}
\begingroup
\setlength{\tabcolsep}{11pt}
\renewcommand{\arraystretch}{1.05}
\resizebox{0.95\linewidth}{!}{
\small
\begin{tabular}{cccc}
\toprule
Method & Size & FID ↓ & IS ↑ \\
\midrule
\textcolor{gray}{Full-Precision} & \textcolor{gray}{208MB} & \textcolor{gray}{2.30} & \textcolor{gray}{279.44} \\
\midrule
\textbf{\spc Baseline} & 52MB & 10.34 & 144.22 \\
\textbf{\spc + HCD} & \textbf{52MB} & \textbf{3.50} & \textbf{247.33} \\
\midrule
\textbf{\pct Baseline} & 52MB & 252.41 & 1.98 \\
\textbf{\pct + SR} & 52MB & 11.34 & 139.77 \\
\textbf{\pct + HCD + SR (HyGenQ)} & \textbf{52MB} & \textbf{3.61} & \textbf{246.00} \\
\bottomrule
\end{tabular}
}
\endgroup
\end{table}

To evaluate the effectiveness of each component in HyGenQ, we conduct an ablation study on the MAR-B model, under a configuration of 64 autoregressive iterations and 100 diffusion timesteps.
HyGenQ consists of two key modules: \textbf{Hierarchical Cluster Decoupling (HCD)} for handling excessive outliers and \textbf{Scaling Recalibration (SR)} for suppressing amplified anomalies.
Results are summarized in Table~\ref{table ablation 1}.
We conduct the ablation under both \spc and \pct settings. 
For the \spc setting, which does not suffer from degradation caused by amplified anomalies, SR is not required, and two method variants are considered in our ablation:
\textbf{(i) \spc Baseline.} This setting keeps the boundary layers in the diffusion procedure in full precision, while applying basic linear quantization to all other linear layers without any mitigation.
\textbf{(ii) \spc Baseline + HCD.} Incorporating HCD alone mitigates excessive outliers and further improves the generative quality.
For the \pct setting, three method variants are considered in our ablation:
\textbf{(i) \pct Baseline.} This setting applies basic linear quantization to all linear layers without any mitigation. In this case, anomalies result in catastrophic model collapse.
\textbf{(ii) \pct Baseline + SR.} Incorporating SR alone effectively suppresses anomalies, thereby avoiding the severe degradation observed in the baseline. As a result, the generative quality is significantly restored.
\textbf{(iii) \pct Baseline + HCD + SR (HyGenQ).} Building upon SR, we further integrate HCD to address the challenge of outliers, forming the complete HyGenQ framework.
These results show that HyGenQ effectively mitigates excessive outliers and amplified anomalies, enabling it to preserve generation quality under aggressive quantization.

\vspace{-2mm}

\begin{table}[t]
\vspace{-4mm}
\centering
\caption{
End-to-end inference efficiency on MAR models with batch size 32.
Spd. denotes latency speedup over full precision, T./Img. denotes the average generation time per image, Param. denotes the storage cost (MB) of the quantization parameters for a single autoregressive iteration, and Calib. denotes the offline calibration time (min).
}
\label{tab:inference_efficiency}
\vspace{-2mm}
\begingroup
\setlength{\tabcolsep}{2.2pt}
\renewcommand{\arraystretch}{1.05}
\resizebox{\linewidth}{!}{
\small
\begin{tabular}{lccc cccc cccc}
\toprule
Size & Method & W/A & \spc Spd. & \spc T./Img. & \spc Param. & \spc Calib. & \pct Spd. & \pct T./Img. & \pct Param. & \pct Calib. \\
\midrule
B & FP     & 32/32 & 1.00$\times$ & 0.979 & -- & -- & 1.00$\times$ & 0.979 & -- & -- \\
B & PTQ4DM & 8/8   & 1.49$\times$ & 0.659 & 0.0175 & 15 & 1.47$\times$ & 0.666 & 0.0191 & 15 \\
B & HyGenQ & 8/8   & 1.47$\times$ & 0.667 & 0.0175 & 22 & 1.44$\times$ & 0.678 & 0.0176 & 22 \\
\midrule
L & FP     & 32/32 & 1.00$\times$ & 1.618 & -- & -- & 1.00$\times$ & 1.618 & -- & -- \\
L & PTQ4DM & 8/8   & 1.73$\times$ & 0.936 & 0.0224 & 20 & 1.72$\times$ & 0.939 & 0.0239 & 20 \\
L & HyGenQ & 8/8   & 1.77$\times$ & 0.915 & 0.0224 & 22 & 1.70$\times$ & 0.953 & 0.0224 & 23 \\
\midrule
H & FP     & 32/32 & 1.00$\times$ & 2.691 & -- & -- & 1.00$\times$ & 2.691 & -- & -- \\
H & PTQ4DM & 8/8   & 2.16$\times$ & 1.247 & 0.0318 & 44 & 2.19$\times$ & 1.231 & 0.0333 & 45 \\
H & HyGenQ & 8/8   & 1.99$\times$ & 1.353 & 0.0318 & 53 & 1.79$\times$ & 1.505 & 0.0318 & 54 \\
\bottomrule
\end{tabular}
}
\endgroup
\vspace{-4mm}
\end{table}

\subsection{Inference Efficiency}

We further evaluate the end-to-end inference efficiency of quantized MAR models across different scales.
Following ViDiT-Q~\cite{zhao2024viditq}, we use cuBLASLt INT8 GEMM through a lightweight wrapper implemented in C++ and CUDA for quantized linear operations.
We offline bake per-(step,$i$) scale/zero-point tables to match the gradually drifting activation distributions across sampling, and capture the forward pass of the diffusion loss network as a CUDA Graph, which is replayed at each step with only its parameters updated to reduce kernel launch overhead during sampling.
All methods are tested under the same sampling configuration.
As shown in Table~\ref{tab:inference_efficiency}, HyGenQ achieves substantial acceleration over full-precision inference under both \spc and \pct settings.
Under \spc, HyGenQ reduces the average generation time to 0.667s, 0.915s, and 1.353s per image on MAR-B, MAR-L, and MAR-H, respectively.
Under \pct, the corresponding times are 0.678s, 0.953s, and 1.505s.
Compared with PTQ4DM, HyGenQ maintains comparable efficiency, with slight extra latency mainly caused by inference-time recalibration, especially on MAR-H where recalibration is triggered more frequently.
Nevertheless, it still achieves clear speedups over full-precision inference while delivering much better generation quality.
We also report the deployment overhead of quantized methods: the quantization parameters required for a single autoregressive iteration occupy only tens of KB across all model scales, and the offline calibration is a one-time cost (up to 54 minutes on MAR-H for HyGenQ) that does not affect online inference; the moderately longer calibration of HyGenQ over PTQ4DM reflects its finer-grained per-(step,$i$) calibration.

\vspace{-3mm}
\subsection{Lower-Bit Quantization}

We further evaluate more aggressive lower-bit quantization settings.
As shown in Table~\ref{tab:w8a6}, under W8A6 quantization with the \spc setting, HyGenQ consistently outperforms SmoothQuant~\cite{xiao2023smoothquant} and PTQ4DM~\cite{shang2023ptq4dm} across all MAR scales.
However, the overall performance still drops sharply compared to W8A8, indicating that hybrid IGMs are highly sensitive to reduced activation precision.
Figure~\ref{lower} further visualizes W4A8 and W4A4 results.
When the bit-width is reduced too aggressively, accumulated errors across autoregressive iterations and diffusion timesteps dominate the generation process, causing severe collapse.
Therefore, excessively low bit-widths fail to support stable hybrid iterative generation.

\begin{table}[t]
\centering
\captionsetup{width=\linewidth}
\caption{W8A6 comparison on MAR with 8,000 generated images under the \spc setting.}
\label{tab:w8a6}
\vspace{-2mm}
\begingroup
\setlength{\tabcolsep}{5pt}
\renewcommand{\arraystretch}{1.05}
\resizebox{0.95\linewidth}{!}{
\begin{tabular}{lcccc}
\toprule
Method & Model & Precision (W/A) & FID $\downarrow$ & IS $\uparrow$ \\
\midrule
\spc SmoothQuant~\cite{xiao2023smoothquant} & MAR-B & 8/6 & 156.01 & 6.77 \\
\spc PTQ4DM~\cite{shang2023ptq4dm} & MAR-B & 8/6 & 299.86 & 1.92 \\
\spc \textbf{HyGenQ (Ours)} & MAR-B & 8/6 & \textbf{144.12} & \textbf{8.07} \\
\midrule
\spc SmoothQuant~\cite{xiao2023smoothquant} & MAR-L & 8/6 & 352.84 & 1.72 \\
\spc PTQ4DM~\cite{shang2023ptq4dm} & MAR-L & 8/6 & 287.09 & 1.75 \\
\spc \textbf{HyGenQ (Ours)} & MAR-L & 8/6 & \textbf{251.34} & \textbf{2.01} \\
\midrule
\spc SmoothQuant~\cite{xiao2023smoothquant} & MAR-H & 8/6 & 368.06 & 1.49 \\
\spc PTQ4DM~\cite{shang2023ptq4dm} & MAR-H & 8/6 & 366.82 & 1.56 \\
\spc \textbf{HyGenQ (Ours)} & MAR-H & 8/6 & \textbf{363.43} & \textbf{1.69} \\
\bottomrule
\end{tabular}
}
\endgroup
\vspace{-3mm}
\end{table}

\begin{table}[t]
\centering
\captionsetup{width=\linewidth}
\caption{Analysis of calibration protocols on MAR-L. The upper block compares global calibration strategies with the adopted per-step setting, while the lower block evaluates the robustness of per-step calibration.}
\label{table:calib_protocol}
\vspace{-2mm}
\begingroup
\setlength{\tabcolsep}{3pt}
\renewcommand{\arraystretch}{1.05}
\resizebox{0.95\linewidth}{!}{
\begin{tabular}{lccccc}
\toprule
Method & Calib. & Samples & Seed & FID $\downarrow$ & IS $\uparrow$ \\
\midrule
HyGenQ & Global Max & 32 & 1 & 8.29 & 169.34 \\
HyGenQ & PTQ4DM-style~\cite{shang2023ptq4dm} & 32 & 1 & 8.51 & 171.24 \\
HyGenQ & PQD-style~\cite{ye2025pqd} & 32 & 1 & 7.93 & 173.63 \\
HyGenQ & Per-step & 32 & 1 & \textbf{6.06} & \textbf{193.94} \\
\midrule
HyGenQ & Per-step & 128 & 1 & 6.36 & 191.32 \\
HyGenQ & Per-step & 64 & 1 & 6.23 & 193.07 \\
HyGenQ & Per-step & 32 & 21 & 6.11 & 192.62 \\
HyGenQ & Per-step & 32 & 42 & 6.09 & 191.34 \\
\bottomrule
\end{tabular}
}
\endgroup
\vspace{-4mm}
\end{table}

\vspace{-2mm}
\subsection{Analysis of Calibration Protocol}
\label{ana_calib}

We further analyze the generation quality of hybrid IGMs when each layer uses only one shared set of quantization parameters.
As shown in Table~\ref{table:calib_protocol}, PTQ4DM-style~\cite{shang2023ptq4dm} and PQD-style~\cite{ye2025pqd} calibration improve over Global Max, but their gains remain limited.
Compared with the per-step setting, global quantizers still lead to a clear degradation.
This indicates that calibration selection alone cannot fully handle the activation shifts across autoregressive iterations and diffusion timesteps.
Therefore, to focus on the core challenges of excessive outliers and amplified anomalies, we assign separate quantization parameters to each autoregressive iteration and diffusion timestep in our main experiments.
We also evaluate the robustness of the adopted per-step setting.
Changing the calibration size from 32 to 64 or 128 and varying the random seed only cause minor performance changes, showing that the calibration protocol is stable.

\subsection{Comparison of HCD Clustering Strategies}

We further compare different grouping strategies for outlier-channel discovery in HCD, including genetic search, $k$-means++~\cite{arthur2007k}, and a threshold-based strategy.
All experiments follow the calibration protocol in Sec.~\ref{ana_calib}, with only the grouping strategy in HCD replaced.

As shown in Table~\ref{table:hcd_grouping_strategies}, $k$-means++ performs worse across model scales, indicating that standard clustering is less effective for separating abnormal channels under highly skewed activation distributions.
The threshold-based strategy achieves competitive FID on MAR-B and MAR-H, but its IS remains lower than genetic search, suggesting that a fixed threshold may suppress or misassign informative outlier channels.
In contrast, genetic search achieves the best IS across all model scales and the best FID on MAR-L, providing a better overall trade-off between distribution fidelity and semantic quality.
These results show that the genetic search used in HCD is more suitable for adaptive outlier-channel grouping in hybrid IGMs.

\begin{table}[t]
\centering
\captionsetup{width=0.95\linewidth}
\caption{Comparison of HCD grouping strategies under the 8,000-image setting on ImageNet \(256\times256\).}
\label{table:hcd_grouping_strategies}
\vspace{-2mm}
\begingroup
\setlength{\tabcolsep}{7.5pt}
\renewcommand{\arraystretch}{1.05}
\resizebox{0.95\linewidth}{!}{
\begin{tabular}{llccc}
\toprule
Method & Model & Clustering strategy & FID $\downarrow$ & IS $\uparrow$ \\
\midrule
HyGenQ & MAR-B & $k$-means++ & 8.17 & 151.45 \\
HyGenQ & MAR-B & threshold ($P_{99}$) & \textbf{6.96} & 175.72 \\
HyGenQ & MAR-B & genetic & 7.14 & \textbf{178.74} \\
\midrule
HyGenQ & MAR-L & $k$-means++ & 6.26 & 187.84 \\
HyGenQ & MAR-L & threshold ($P_{99}$) & 6.11 & 188.66 \\
HyGenQ & MAR-L & genetic & \textbf{6.06} & \textbf{193.94} \\
\midrule
HyGenQ & MAR-H & $k$-means++ & 7.31 & 182.37 \\
HyGenQ & MAR-H & threshold ($P_{99}$) & \textbf{6.99} & 185.73 \\
HyGenQ & MAR-H & genetic & 7.19 & \textbf{188.32} \\
\bottomrule
\end{tabular}
}
\endgroup
\vspace{-3mm}
\end{table}

\vspace{-2mm}
\subsection{Hyperparameter Sensitivity}\label{sec:Hyperparameter Sensitivity}

\begin{table}[t]
\centering
\captionsetup{width=1\linewidth}
\caption{Sensitivity of cluster number ranges in HCD.}
\label{tab:hcd-k}
\begingroup
\setlength{\tabcolsep}{10pt}
\renewcommand{\arraystretch}{1.05}
\resizebox{0.95\linewidth}{!}{
\begin{tabular}{cccc}
\toprule
Method & Range(k) & FID $\downarrow$ & IS $\uparrow$ \\
\midrule
\textcolor{gray}{FP} & \textcolor{gray}{-} & \textcolor{gray}{5.1166} & \textcolor{gray}{207.2319} \\
\midrule
PTQ4DM & - & 9.2551 & 132.0545 \\
\midrule
HyGenQ & (2,10) & 6.0746 & 192.5079 \\
HyGenQ & (2,15) & 6.0746 & 192.5079 \\
\textbf{HyGenQ (Ours)} & \textbf{(2,20)} & \textbf{6.0596} & \textbf{193.9367} \\
HyGenQ & (2,30) & 6.0746 & 192.5079 \\
HyGenQ & (2,40) & 6.5120 & 184.8754 \\
HyGenQ & (2,50) & 6.5120 & 184.8754 \\
\bottomrule
\end{tabular}
}
\endgroup
\vspace{-4mm}
\end{table}

\begin{table}[t]
\centering
\captionsetup{width=1\linewidth}
\caption{Sensitivity of $\omega$ in HCD.}
\label{tab:hcd-omega}
\vspace{-2mm}
\begingroup
\setlength{\tabcolsep}{10pt}
\renewcommand{\arraystretch}{1.05}
\resizebox{0.95\linewidth}{!}{
\begin{tabular}{cccc}
\toprule
Method & $\omega$ & FID $\downarrow$ & IS $\uparrow$ \\
\midrule
\textcolor{gray}{FP} & \textcolor{gray}{-} & \textcolor{gray}{5.1166} & \textcolor{gray}{207.2319} \\
\midrule
PTQ4DM & - & 9.2551 & 132.0545 \\
\midrule
HyGenQ & 0.0001 & 6.0976 & 194.4706 \\
\textbf{HyGenQ (Ours)} & \textbf{0.001} & \textbf{6.0596} & \textbf{193.9367} \\
HyGenQ & 0.005 & 6.0746 & 192.5079 \\
HyGenQ & 0.01 & 6.0746 & 192.5079 \\
HyGenQ & 0.05 & 6.1136 & 191.6326 \\
HyGenQ & 0.1 & 6.0617 & 193.4275 \\
HyGenQ & 0.5 & 6.1532 & 191.7512 \\
\bottomrule
\end{tabular}
}
\endgroup
\vspace{-2mm}
\end{table}

\begin{table}[t]
\centering
\captionsetup{width=0.95\linewidth}
\caption{Robustness to GA hyperparameters in HCD.}
\label{tab:ga-robustness}
\vspace{-2mm}
\begingroup
\setlength{\tabcolsep}{6pt}
\renewcommand{\arraystretch}{1.05}
\resizebox{0.95\linewidth}{!}{
\begin{tabular}{ccccccc}
\toprule
Method & \(N_{pop}\) & T & \(P_c\) & \(P_m\) & FID $\downarrow$ & IS $\uparrow$ \\
\midrule
\textcolor{gray}{FP} & \multicolumn{4}{c}{\textcolor{gray}{---}} & \textcolor{gray}{5.1166} & \textcolor{gray}{207.2319} \\
\midrule
PTQ4DM & \multicolumn{4}{c}{---} & 9.2551 & 132.0545 \\
\midrule
\textbf{HyGenQ} & \textbf{50} & \textbf{20} & \textbf{0.5} & \textbf{0.2} & \textbf{6.0596} & \textbf{193.9367} \\
HyGenQ & \textbf{20} & 20 & 0.5 & 0.2 & 6.4661 & 192.1524 \\
HyGenQ & \textbf{60} & 20 & 0.5 & 0.2 & 6.4661 & 192.1524 \\
HyGenQ & 50 & \textbf{10} & 0.5 & 0.2 & 6.6079 & 182.7686 \\
HyGenQ & 50 & \textbf{50} & 0.5 & 0.2 & 6.6079 & 182.7686 \\
HyGenQ & 50 & 20 & \textbf{0.2} & 0.2 & 6.3359 & 188.0015 \\
HyGenQ & 50 & 20 & \textbf{0.7} & 0.2 & 6.6079 & 182.7686 \\
HyGenQ & 50 & 20 & 0.5 & \textbf{0.1} & 6.6079 & 182.7686 \\
HyGenQ & 50 & 20 & 0.5 & \textbf{0.5} & 6.5201 & 191.1514 \\
\bottomrule
\end{tabular}
}
\endgroup
\vspace{-2mm}
\end{table}

To validate the robustness and practicality of our method, we conduct controlled ablations on key hyperparameters using MAR-L on ImageNet $256\times256$~\cite{deng2009imagenet} with 8,000 generated samples.
The results are summarized as follows.

\paragraph{\textbf{Sensitivity of the Number of Clusters \(k\) in HCD}}
Our method is partly motivated by its ability to avoid manual tuning of the cluster number $k$, a common issue in K-Means.
As shown in Table~\ref{tab:hcd-k}, our method remains stable across different cluster-number ranges.
Although we set \(k{=}20\) as the default upper bound, nearby ranges produce similar performance, indicating that our method does not require fine-grained tuning of \(k\).

We further evaluate the stability of outlier channels detected by HCD across random seeds and input classes with MAR-B.
For random seeds, we define the model-seed set as \(\mathcal{M}=\{0,1,2\}\), and compute the IoU, denoted as \({\mathrm{IoU}}(S_a,S_b)\), between the detected outlier-channel sets for any \(a,b\in\mathcal{M}\) and \(a\neq b\).
For input classes, we construct different calibration sequences through three random samplings.
To avoid interference from neutral channels that may fluctuate across inputs, we use the retention metric \(\mathrm{Ret}(S_{\mathrm{core}},S_i)=|S_{\mathrm{core}}\cap S_i|/|S_{\mathrm{core}}|\), where \(S_{\mathrm{core}}\) denotes the core anomalous-channel set and \(S_i\) denotes the detected outlier-channel set under the \(i\)-th calibration split.
As shown in Table~\ref{tab:hcd-stability}, both metrics remain 1.000, indicating that the outlier channels detected by HCD are consistent across random seeds and different input classes.

\begin{table}[t]
\centering
\caption{Stability of HCD-detected outlier channels on MAR-B. Here, \(S_a,S_b\) are detected sets under different seeds; \(S_i\) is the detected set under the \(i\)-th input-class split; and \(S_{\mathrm{core}}\) is the core anomalous-channel set.}
\label{tab:hcd-stability}
\vspace{-2mm}
\begingroup
\setlength{\tabcolsep}{5pt}
\renewcommand{\arraystretch}{1.05}
\resizebox{0.95\linewidth}{!}{
\small
\begin{tabular}{ccccc}
\toprule
Metric & Scope & Min & Max & Mean \\
\midrule
\({\mathrm{IoU}}(S_a,S_b)\) & Random seeds & 1.000 & 1.000 & 1.000 \\
\({\mathrm{Ret}}(S_{\mathrm{core}},S_i)\) & Input classes & 1.000 & 1.000 & 1.000 \\
\bottomrule
\end{tabular}
}
\endgroup
\vspace{-2mm}
\end{table}

\paragraph{\textbf{Sensitivity of the Weighting Factor \(\omega\) in HCD}}
The weighting factor $\omega$ in our method is introduced to emphasize abnormal channels while downweighting the influence of normal channels during clustering.
Table~\ref{tab:hcd-omega} evaluates the effect of \(\omega\), which balances abnormal and normal channels during clustering.
The results remain stable across a wide range of small \(\omega\) values, suggesting that our method is not sensitive to this hyperparameter.
We use \(\omega{=}0.001\) as the default setting.

\paragraph{\textbf{Robustness of Genetic Algorithm Parameters in HCD}}
We use a standard Genetic Algorithm following the DEAP framework~\cite{fortin2012deap}.
As shown in Table~\ref{tab:ga-robustness}, the default setting achieves the best trade-off between performance and stability.
Although alternative GA configurations slightly degrade performance, they still clearly outperform PTQ4DM~\cite{shang2023ptq4dm}, indicating that our method is robust to fine-grained GA parameter choices.

\paragraph{\textbf{Sensitivity of the Gaussian Bound in SR}}
In our SR, the Gaussian Bound \(\Phi^{-1}(\eta)\) is derived from the activation distribution of the FP model.
We first perform a KS analysis on the boundary-layer activations, where the mean KS value is only \(0.0211\) and the maximum value is \(0.0617\), indicating that the Gaussian approximation provides a reasonable reference.
Therefore, we compute the bound under the standard Gaussian assumption to simplify the bound estimation.
Taking MAR-B with a batch size of 32 as the reference setting, one forward pass through the boundary layers produces \(32\times (16+1024) =33{,}280\) activation values, which is conservatively approximated as \(N=3.4\times10^4\).
We require these values to be covered by \([-G,G]\) with probability \(0.999\), namely \((2\Phi(G)-1)^N=0.999\).
This yields \(\eta=0.9999999853\) and \(G=\Phi^{-1}(\eta)=5.544815\), which is rounded to \(5.545\).
We therefore set the Gaussian Bound to \([-5.545,5.545]\).
To avoid tuning the bound separately for different model variants, layers, and generation settings, we use this bound globally across all experiments.
As shown in Table~\ref{tab:sr-gb}, \(G=5.545\) achieves stable FID and IS, confirming that this setting is effective and robust.

\section{Open Issues and Potential Insights}
\label{Discussion}

\textbf{The extreme kurtosis of outliers.} As discussed above, Excessive Outliers challenge traditional quantization techniques.
To tackle this, our method isolates both normal values and outliers as much as possible, offering an effective solution.
However, this strategy introduces additional inference latency, especially on MAR-H, where the extra SR operations noticeably weaken the acceleration gains. Therefore, reducing this overhead remains an important direction for future improvement.

\textbf{The exceedance of PTQ's applicability range.} The large fluctuations of Amplified Anomalies during calibration and inference complicate the calibration process in the PTQ framework, rendering it largely ineffective.
To address this, we strike a balance between the constraints of the PTQ framework and the need to mitigate mismatches between calibration and inference.
However, this approach comes with some computational overhead, which, although negligible, still requires more consideration for optimization.

\textbf{The severe accumulation of errors.}  As can be seen from the experimental results (Table~\ref{table1} and Table~\ref{table2}), due to the severity of anomalies, while our method successfully prevents the model collapse observed in the baseline methods under \pct-versions, it still demonstrates performance degradation compared to \spc-versions, particularly in larger, deeper, and more iterative models such as MAR-H. This highlights the trade-off between the loss in generative performance due to quantization and the resulting improvement in inference speed. In practical deployment, this could significantly reduce the usability and effectiveness of the MAR-H model under \pct-versions. Therefore, we believe that the issue of anomalies within the PTQ framework still warrants further investigation.

\begin{table}[t]
\centering
\caption{Sensitivity of Gaussian Bound (G) in SR.}
\label{tab:sr-gb}
\vspace{-2mm}
\begingroup
\setlength{\tabcolsep}{10pt}
\renewcommand{\arraystretch}{1.05}
\resizebox{0.95\linewidth}{!}{
\vspace{-2mm}
\begin{tabular}{cccc}
\toprule
Method & G & FID $\downarrow$ & IS $\uparrow$ \\
\midrule
\textcolor{gray}{FP} & \textcolor{gray}{-} & \textcolor{gray}{5.1166} & \textcolor{gray}{207.2319} \\
\midrule
PTQ4DM & - & 9.2551 & 132.0545 \\
\midrule
HyGenQ & 3 & 6.6878 & 183.5942 \\
HyGenQ & 4 & 6.6113 & 182.9316 \\
HyGenQ & 5 & 6.6691 & 183.7552 \\
\textbf{HyGenQ (Ours)} & \textbf{5.545} & \textbf{6.6079} & \textbf{182.7686} \\
HyGenQ & 6 & 6.6371 & 182.2333 \\
HyGenQ & 7 & 8.1102 & 165.5975 \\
\bottomrule
\end{tabular}
}
\endgroup
\vspace{-4mm}
\end{table}

\vspace{-2mm}
\section{Conclusion}
\label{conclusion}

This paper proposes HyGenQ, a post-training quantization framework for hybrid iterative generative models, targeting two key challenges in multi-step generation: Excessive Outliers and Amplified Anomalies.
Using MAR as a representative autoregressive-diffusion hybrid model, we develop a joint solution with Hierarchical Cluster Decoupling and Scaling Recalibration.
Hierarchical Cluster Decoupling separates stable outlier channels from normal channels and quantizes them independently, preserving both normal-value precision and outlier information.
Scaling Recalibration constrains unpredictable amplified anomalies with a Gaussian Bound and rescales anomalous channels during denoising, preventing severe generation collapse.
Extensive experiments show that HyGenQ achieves high-fidelity W8A8 quantization and consistently outperforms strong baselines under identical iterative generation settings.

In summary, this paper has two primary objectives.
First, we highlight the severe quantization challenges in hybrid iterative generative models, especially excessive outliers and amplified anomalies that become more pronounced in iterative paradigms.
Second, we hope this work can encourage further exploration of quantization for complex generative models and inspire more effective solutions for these challenges.

\bibliography{tcsvt2026}
\bibliographystyle{IEEEtran}

\vfill

\end{document}